\documentclass[twoside,11pt]{article}
\usepackage{showlabels}
\usepackage{jmlr2e}
\usepackage{amsfonts,amsmath,amssymb,amsthm}
\usepackage{enumerate,bbm,ifthen,color}
\usepackage{hyperref,graphicx}
\usepackage[latin1]{inputenc}
\usepackage{natbib}
\def \n1{{n_1}}

\def \la{{\lambda}}

\def \al{\boldsymbol{\alpha}}

\def \H{{\mathcal H}}

\def \E{{\mathbb E}}
\def \P{{\mathbb P}}
\def \Q{\mathbb{Q}}

\def \S{{\mathbf S}}
\def \hS {\hat{\S}}
\def \hdelta {\hat{\delta}}
\def \n{{\mathbf n}}
\def \S{ \Sigma }

\def\as{\ensuremath{\mathrm{a.s}}}

\newcommand{\PVar}{\mathrm{Var}}
\newcommand{\PCov}{\mathrm{Cov}}

\newcommand{\eqsp}{\;}
\newcommand\pscal[2]{ \left\langle #1 , #2 \right\rangle }
\newcommand\chisqdiv[2]{ \ensuremath{ D_{\chi^2}\left( #1 \parallel #2 \right) } }
\newcommand{\id}{I}
\newcommand{\Kb}{\mathbf{K}}
\newcommand{\Pb}{\mathbf{P}}

\newcommand{\Nb}{\mathbf{N}}

\newcommand{\ab}{\mathbf{a}}
\newcommand{\mb}{\mathbf{m}}

\newcommand{\Gb}{\mathbf{G}}
\newcommand{\x}{X}
\newcommand{\condgan}{\ensuremath{\ga_n + d_2^{-1}(\S_1,\ga_n) d_1(\S_1,\ga_n) \ga_n^{-1}n^{-1/2} \to 0}}

\newcommand{\condgaf}{\ensuremath{\gamma_n \equiv \gamma}}

\newcommand{\eqdef}{\ensuremath{\stackrel{\mathrm{def}}{=}}}
\newcommand{\mysec}[1]{Section~\ref{sec:#1}}

\def\max{\operatornamewithlimits{max}}

\def\rmi{\mathrm{i}}
\def\rme{\mathrm{e}}
\def\trace{\mathrm{Tr}}

\def\Id{\mathrm{I}}

\newtheorem{theo}{Theorem}

\newtheorem{prop}[theo]{Proposition}
\newtheorem{lem}[theo]{Lemma}
\newtheorem{cor}[theo]{Corollary}
\newtheorem{remark}[theo]{Remark}

\newcommand{\hsnorm}[1]{\left\Vert #1 \right\Vert_{\mathrm{HS}}}
\newcommand{\carlnorm}[1]{\left\Vert #1 \right\Vert_{\mathcal{C}_1}}
\newcommand{\openorm}[1]{\left\Vert #1 \right\Vert}
\newcommand{\norm}[1]{\left\Vert #1 \right\Vert}

\newcounter{hypA}
\newenvironment{hypA}{\refstepcounter{hypA}\begin{itemize}
  \item[({\bf A\arabic{hypA}})]}{\end{itemize}}
  \newcounter{hypB}
\newenvironment{hypB}{\refstepcounter{hypB}\begin{itemize}
  \item[({\bf B\arabic{hypB}})]}{\end{itemize}}
  \newcounter{hypC}
\newenvironment{hypC}{\refstepcounter{hypC}\begin{itemize}
  \item[({\bf C})]}{\end{itemize}}
    \newcounter{hypD}

\newcommand{\dlim}{\ensuremath{\stackrel{\mathcal{D}}{\longrightarrow}}}
\newcommand{\plim}{\ensuremath{\stackrel{\mathrm{P}}{\longrightarrow}}}

\newcommand{\gauss}{\ensuremath{\mathcal{N}}}

\def\mcf{\mathcal{F}}
\newcommand{\CPE}[3][]
{\ifthenelse{\equal{#1}{}}{\mathbb{E}\left[\left. #2 \, \right| #3 \right]}{\mathbb{E}_{#1}\left[\left. #2 \, \right | #3 \right]}}

\newcommand{\wrt}{with respect to}

\def \HH{{\textbf{H}_0}}
\def \hh{{\textbf{H}_A }}
\def \hhn{{\textbf{H}_A^{n}}}
\def \H{{\mathcal H}}

\def\ltwo{L^2}
\def\lone{L^1}

\def\eg{\textit{e.g.}}

\newcommand{\BEAS}{\begin{eqnarray*}}
\newcommand{\EEAS}{\end{eqnarray*}}
\newcommand{\BEA}{\begin{eqnarray}}
\newcommand{\EEA}{\end{eqnarray}}
\newcommand{\BEQ}{\begin{equation}}
\newcommand{\EEQ}{\end{equation}}
\newcommand{\BAS}{\begin{align*}}
\newcommand{\EAS}{\end{align*}}
\newcommand{\BA}{\begin{align}}
\newcommand{\EA}{\end{align}}
\newcommand{\rb}{\mathbb{R}}

\def\Xset{\mathcal{X}}

\def\Xsigma{\mathfrak{X}}

\newcommand\RKHS[1]{\mathcal{H}}
\def\Rset{\mathbb{R}}
\def\ie{\textit{i.e.}}

\newcommand{\hla}{\hat{\la}}%lambda "hat"
\newcommand{\hT}{\widehat{T}}%T "hat"
\newcommand{\hV}{\hat{V}}%V "hat"
\newcommand{\tT}{\tilde{T}}%T "tilde"
\newcommand{\hmu}{\hat{\mu}}%mu "hat"
\newcommand{\ce}{\overline{e}}

\newcommand{\kmax}{|k|_{\infty}}
\newcommand{\ga}{\gamma}

\newcommand{\sump}{\sum_{p=1}^{\infty}}%sum on q
\newcommand{\sumi}{\sum_{i=1}^{n}}%sum for i=1,...,n
\newcommand{\sumpaq}{\sum_{p,q=1}^\infty}
\newcommand{\sumpq}{\sum_{p \neq q}}%sum on p and q, outside diagonal
\newcommand{\sqfacone}{\frac{n_2}{n_1 n}}%square of factor of sample (1)
\newcommand{\sqfactwo}{\frac{n_1}{n_2 n}}%square of factor of sample (2)
\newcommand{\pons}{\frac{n_1 n_2}{n}}%product on sum

\newcommand{\dan}{\delta_{n}}
\newcommand{\dep}{\pscal{\dan}{e_p}}
\newcommand{\deq}{\pscal{\dan}{e_q}}

\newcommand{\epp}{\varepsilon_{p,p}}
\newcommand{\epq}{\varepsilon_{p,q}}

\newcommand{\meanshift}{\Delta}
\newcommand{\KFDA}{\textsf{KFDA}}
\newcommand{\MMD}{\textsf{MMD}}

\ShortHeadings{Testing for Homogeneity with KFDA}{ }
\firstpageno{1}

\begin{document}
\title{Testing for Homogeneity\\ with Kernel Fisher Discriminant Analysis}

\author{\name Za\"{i}d Harchaoui \email zaid.harchaoui@enst.fr \\
       \addr LTCI, TELECOM ParisTech \& CNRS \\
        46, rue Barrault\\
     75634 Paris cedex 13,  France
       \AND
    \name Francis R. Bach \email francis.bach@mines.org \\
       \addr INRIA - Willow Project-Team \\
       Laboratoire d'Informatique de l'\'{E}cole Normale Sup\'{e}rieure \\
       45, rue d'Ulm \\
        75230 Paris, France
       \AND
       \name \'{E}ric Moulines \email eric.moulines@enst.fr \\
       \addr LTCI, TELECOM ParisTech \& CNRS \\
        46, rue Barrault\\
     75634 Paris cedex 13,  France}
\editor{}
\maketitle

\begin{abstract}
We propose to investigate test statistics
for testing homogeneity in reproducing kernel Hilbert spaces.
Asymptotic null distributions under null hypothesis are derived,
and consistency under fixed and local alternatives is assessed.
Finally, experimental evidence of the performance of the proposed approach
on both artificial data and a speaker verification task is provided.
\end{abstract}

{}

 \section{Introduction}
An important problem in statistics and machine learning consists in testing whether the distributions
of two random variables are identical under the alternative that they may differ in some ways.
More precisely, let $\{X_1^{(1)}, \dots, X_{n_1}^{(1)}\}$ and $\{ X_1^{(2)}, \dots, X_{n_2}^{(2)} \}$ be independent random variables
taking values in an arbitrary input space $\Xset$, with common distributions $\P_1$ and $\P_2$,
respectively. The problem consists in testing the null hypothesis of homogeneity $\HH: \P_1= \P_2$ ,
against the alternative $\hh: \P_1 \ne \P_2$.
This problem arises in many applications, ranging from computational anatomy~\citep{Grenander:Miller:2007}
to speaker segmentation~\citep{Reynolds:2004}. We shall allow the input space $\Xset$ to be quite general,
including for example finite-dimensional euclidean spaces but also function spaces, or more sophisticated 
structures such as strings or graphs \citep[see][]{Shawe:Cristianini:2004} arising in applications such as bioinformatics \citep[see recently][]{Borgwardt:Gretton:Rasch:Kriegel:Schoelkopf:Smola:2006}.

Traditional approaches to this problem are based on cumulative distribution functions (cdf),
and use a certain distance between the empirical cdf obtained from the two samples.
Popular procedures are the two-sample Kolmogorov-Smirnov tests or the Cramer-Von Mises tests~\citep{Lehmann:Romano:2005},
that have been frequently used to address these issues, at least for low-dimensional data.
Although these tests are popular due to their simplicity, they are known to be insensitive to certain characteristics of the distributions, such as densities containing high-frequency components or local features such as narrow bumps.
The low-power of the traditional cdf-based test statistics can be improved on by using test statistics based on probability density estimators. Tests based on kernel density estimators have been studied
by~\citet{Anderson:Hall:Titterington:1994} and \citet{Allen:1997}, using respectively the $\ltwo$ and $\lone$ distances between densities. More recently, the use of wavelet estimators has been proposed and thoroughly analyzed. Adaptive versions of these tests,
that is where smoothing parameters for the density estimator are obtained from the data, 
have been considered by~\citet{Butucea:Tribouley:2006}.

Recently, \cite{Gretton:Borgwardt:Rasch:Schoelkopf:Smola:2006} cast the two-sample homogeneity test
in a kernel-based framework, and have shown that their test statistics, coined Maximum Mean Discrepancy (\MMD) yields as a particular case the $L^2$-distance between kernel density estimators. We propose here to further enhance such an approach by directly incorporating
the covariance structure of the probability distributions into our test statistics, yielding in some sense
to a chi-square divergence between the two distributions. For discrete distributions, it is well-known that
such a normalization yield test statistics with greater power~\citep{Lehmann:Romano:2005}.

The paper is organized as follows. In \mysec{definitions-statements} and
\mysec{homogeneity}, we
state the main definitions and we build our test statistics upon kernel Fisher discriminant analysis. In \mysec{results},
we give the asymptotic distribution of our test statistic under the null hypothesis,
and establish the consistency and the limiting distribution of the test for both fixed and a class of local alternatives. In \mysec{discussion}, we first investigate the limiting power of our test statistics against directional then non-directional sequences of local alternatives in a particular setting, that is when $\P_1$ is the uniform distribution and $\P_2$ is a one-frequency contamination of $\P_1$ on the Fourier basis and the reproducing kernel belongs to the class of periodic spline kernels, and then compare our test statistics with the \MMD~test statistics in terms of limiting power. In Section \ref{sec:experiments} we provide experimental evidence of the performance
of our test statistic on a speaker identification task.
Detailed proofs are presented in the last sections.

\section{Mean and covariance in reproducing kernel Hilbert spaces}
\label{sec:definitions-statements}

We first highlight the main assumptions on the reproducing kernel,
and then introduce operator-theoretic tools for defining the mean element and the covariance operator
associated with a reproducing kernel.

\subsection{Reproducing kernel Hilbert spaces}
Let $(\Xset,d)$ be a separable measurable metric space, and denote by $\Xsigma$ the associated $\sigma$-algebra.
Let $X$  be $\Xset$-valued random variable, with probability measure $\P$, and the expectation with respect to $\P$
is denoted by $\E$.
Consider a Hilbert space $(\H,\pscal{\cdot}{\cdot}_{{\H}})$ of functions from $\Xset$ to $\Rset$.  The
Hilbert space ${\H}$ is a reproducing kernel Hilbert space (RKHS) if at each $x \in \Xset$, the point evaluation operator $\delta_x : {\H} \to \Rset$,
which maps $f \in {\H}$ to $f (x) \in \Rset$, is a bounded linear functional. To each point $x \in \Xset$,
there corresponds an element $\Phi(x) \in  {\H}$ such that
$\pscal{\Phi(x)}{f}_{{\H}} = f(x)$ for all $f \in {\H}$,  and $\pscal{\Phi(x)}{\Phi(y)}_{{\H}}= k(x, y)$,
where $k : \Xset \times \Xset \to \Rset$ is a positive definite kernel~\citep{Aronszajn1950Theory}.
In this situation, $\Phi(\cdot)$ is the Aronszajn-map, and we denote by
$\norm{f}_{\H}= \pscal{f}{f}_{{\H}}^{1/2}$ the associated norm.
It is assumed from now on that $\H$ is a separable Hilbert space. Note that this is
always the case if $\Xset$ is a separable metric space and if the kernel is continuous
\citep[see][]{Steinwart:Hush:Scovel:2006a}.
We make the following two assumptions on the kernel:
\begin{hypA}
\label{hypA:bounded_kernel}
The kernel $k$ is bounded, that is $|k|_{\infty}\eqdef \sup_{(x,y) \in \Xset \times \Xset} k(x,y) < \infty$.
\end{hypA}
\begin{hypA}
\label{hypA:Kernel:injection} For all probability distributions $\P$ on $(\Xset,\Xsigma)$,
the RKHS associated with $k(\cdot,\cdot)$ is dense in $\ltwo(\P)$.
\end{hypA}
Note that some of our results (such as the limiting distribution under the null distribution) are valid without assumption (A\ref{hypA:Kernel:injection}), while consistency results against fixed or local alternatives do need (A\ref{hypA:Kernel:injection}).
Assumption (A\ref{hypA:Kernel:injection}) is true in particular for the gaussian kernel on $\Rset^d$
as shown in~\citep[Theorem 2]{Steinwart:Hush:Scovel:2006b},
and that $\Xset$ may be a discrete space~\citep[Corollary 3]{Steinwart:Hush:Scovel:2006b}.

\subsection{Mean element and covariance operator}
We shall need some operator-theoretic tools~\citep[see][]{Aubin:2000}, to define mean elements and covariance operators.
A linear operator $T$ is said to be \textit{bounded} if there is a number $C$ such that $\norm{Tf}_{\H} \leq C \norm{f}_{\H}$ for all $f \in \H$.
The operator-norm of $T$ is then defined as the minimum of such numbers $C$, that is $\openorm{T} = \sup_{\norm{f}_{\H} \leq 1} \norm{Tf}_{\H}$. Furthermore, a bounded linear operator $T$ is said to be Hilbert-Schmidt, if the Hilbert-Schmidt-norm $\hsnorm{T}=\{\sum_{p=1}^{\infty} \pscal{T e_p}{T e_p}_{{\H}} \}^{1/2}$ is finite, where $\{e_p \}_{p \geq 1}$ is any complete orthonormal basis of $\H$. Note that $\hsnorm{T}$ is independent of the choice of the orthonormal basis. We shall make frequent use of tensor product notations. The tensor product operator $ u \otimes v $ for $u,v \in \H$ is defined
for all $f \in \H$ as $(u \otimes v)f = \pscal{v}{f}_{\H} u$.

We now introduce the mean element and covariance operator~\citep[see][]{Blanchard:Bousquet:Zwald:2007}. If $\int k^{1/2}(x,x) \P(dx) < \infty$, the mean element $\mu_\P$ is defined as the unique element in $\H{}$ satisfying for all functions $f \in \H$,
\begin{equation}
\label{eq:definition-moyenne}
\pscal{\mu_{\P}}{f}_{{\H}} = \P f \eqdef \int f d\P \eqsp.
\end{equation}
If furthermore $\int  k(x,x) \P(dx) < \infty$,
then the covariance operator $\S_\P$ is defined as the unique linear operator onto $\H{}$ satisfying
for all $f,g \in \H$,
\begin{equation}
\label{eq:definition-covariance}
\pscal{f}{\S_\P g}_{{\H}} \eqdef  \int (f - \P f) (g - \P g) d \P \eqsp,
\end{equation}
that is $\pscal{f}{\S_\P g}_{{\H}}$ is the covariance between $f(X)$ and $g(X)$ where $X$ is distributed according to $\P$.
Note that the mean element and covariance operator are well-defined when (A\ref{hypA:bounded_kernel}) is satisfied.
Moreover, when assumption (A\ref{hypA:Kernel:injection}) is satisfied, then the map from $\P \mapsto \mu_\P$ is injective.
Note also that the operator  $\S_\P$ is a self-adjoint nonnegative trace-class operator.
In the sequel, the dependence of $\mu_\P$ and $\Sigma_\P$ in $\P$ is omitted whenever there is no risk of confusion.

We now define what we later denote by $\S^{-1/2}$ in our proofs. For
a compact operator $\S$, the range $\mathcal{R}(\S^{1/2})$ of $\S^{1/2}$ is defined as 
$\mathcal{R}(\S^{1/2})=\{\S^{1/2}f, \; f\in\H \}$, and may be characterized by 
$\mathcal{R}(\S^{1/2})=\{f\in\H, \; \sum_{p=1}^{\infty}\la_p \pscal{f}{e_p}_{\H}^2 < \infty,\; f \perp \mathcal{N}(\S^{1/2}) \}$,
where $\{\la_p, e_p \}_{p \geq 1}$ are the nonzero eigenvalues and eigenvectors of $\S$, 
and $\mathcal{N}(\S) =\{f \in \H, \; \S f = 0 \}$ is the null-space of $\S$, that is functions which are constant in the support of $\P$. Defining 
$\mathcal{R}^{-1}(\S^{1/2})= \{g\in\H, \; g=\sum_{p=1}^{\infty}\la_p^{-1/2} \pscal{f}{e_p}_{\H} e_p,\; f \in \mathcal{R}(\S^{1/2}) \}$, we observe that $\S^{1/2}$ is a one-to-one mapping between $\mathcal{R}^{-1}(\S^{1/2})$
and $\mathcal{R}(\S^{1/2})$. Thus, restricting the domain of $\S^{1/2}$ to $\mathcal{R}^{-1}(\S^{1/2})$,
we may define its inverse for all $f \in \mathcal{R}(\S^{1/2})$ as $\S^{-1/2}f = \sum_{p=1}^{\infty}\la_p^{-1/2} \pscal{f}{e_p}_{\H} e_p$.
The null space may be reduced to the null element (in particular for the gaussian kernel), or may be infinite-dimensional.
Similarly, there may be infinitely many strictly positive eigenvalues (true nonparametric case) or finitely many
(underlying finite-dimensional problems).

Given a sample $\{X_1, \dots, X_n \}$, the empirical estimates respectively of the mean element and the covariance operator are then defined as follows:
\begin{align}
\label{eq:def_mean_cov}
\hat{\mu} &\eqdef n^{-1} \sum_{i=1}^{n} k(X_i, \cdot) \eqsp ,\\
\label{eq:def_mean_cov2}\hat{\S} &\eqdef n^{-1} \sum_{i=1}^{n} k(X_i, \cdot)  \otimes k(X_i, \cdot) - \hat{\mu}\otimes\hat{\mu}
\eqsp .
\end{align}
By the reproducing property, they lead, on the one hand, to empirical means as from~(\ref{eq:def_mean_cov}) we have $\pscal{\hmu}{f}=n^{-1} \sum_{i=1}^{n} f(X_i)$ for all $f \in \H$, and on the other hand, to empirical covariances as from~(\ref{eq:def_mean_cov2}) we have $\langle f, \hat{\Sigma} g \rangle_{\H} = n^{-1} \sum_{i=1}^{n} f(X_i) g(X_i)
- \{ n^{-1} \sum_{i=1}^{n} f(X_i) \}\{n^{-1} \sum_{i=1}^{n} g(X_i) \}$ for all $f,g \in \H$.

\section{KFDA-based test statistic}

\label{sec:homogeneity}
Our two-sample homogeneity test can be formulated as follows.
Let $\{X_1^{(1)},\dots,X_{n_1}^{(1)}\}$ and
$\{X_1^{(2)},\dots,X_{n_2}^{(2)}\}$
two independent identically distributed samples (iid) respectively from $\P_1$ and $\P_2$,
having mean and covariance operators given by $(\mu_{1},\S_{1})$ and $(\mu_{2},\S_{2})$.
We build our test statistics using a (regularized) kernelized version of the Fisher discriminant analysis.
Denote by $\S_{W} \eqdef (n_1/n) \S_{1} + (n_2/n) \S_{2}$ the pooled covariance operator, where $n \eqdef n_1+n_2$,
corresponding to the within-class covariance matrix in the finite-dimensional setting \citep[see][]{Hastie:Tibshirani:Friedman:2001}.

\subsection{Maximum Kernel Fisher Discriminant Ratio}

Let us denote $\S_{B} \eqdef (n_1 n_2/n^2) (\mu_{2}-\mu_{1})
\otimes (\mu_{2}-\mu_{1})$  the between-class covariance operator.
For $a=1,2$, denote by  $(\hat{\mu}_{a},\hat{\S}_{a})$ respectively the
empirical estimates of the mean element and the covariance operator, defined as previously stated in (\ref{eq:def_mean_cov}) and  (\ref{eq:def_mean_cov2}).
Denote  $\hat{\S}_{W} \eqdef (n_1/n) \hat{\S}_1 + (n_2/n) \hat{\S}_2$ the empirical pooled covariance estimator,
and $\hat{\S}_{B} \eqdef (n_1 n_2/n^2) (\hat{\mu}_{2}-\hat{\mu}_{1}) \otimes (\hat{\mu}_{2}-\hat{\mu}_{1})$ the empirical
between-class covariance operator. Let $\{\gamma_n\}_{n \geq 0}$ be a sequence of strictly positive
numbers. The \emph{maximum kernel Fisher discriminant ratio} serves as a basis of our test statistics:
\begin{equation}
n \max_{f \in \H}
\frac{\pscal{f}{\hat{\S}_{B}f}_{\H}}{\pscal{f}{ (\hS_W + \gamma_n \Id) f}_{\H}}
= n_1 n_2/n \; \norm{(\hS_{W} +\gamma_n \Id)^{-1/2} (\hmu_2 - \hmu_1) }_{\H}^2 \eqsp,
\end{equation}
where $\Id$ denotes the identity operator. Note that if the input space is Euclidean, \eg, $\Xset= \Rset^d$,  the kernel is
linear $k(x,y)= x^{T}y$ and $\ga_n =0$, this quantity matches the so-called Hotelling's $T^2$-statistic in the two-sample case~\citep{Lehmann:Romano:2005}. 

We shall make the following assumptions respectively on $\S_1$ and $\S_2$
\begin{hypB}
\label{hypB:sq_sum_eigs}
For $u=1,2$, the eigenvalues $\{\la_p(\S_u)\}_{p \geq 1}$
satisfy $\sum_{p=1}^{\infty} \la_p^{1/2}(\S_u) < \infty$.
\end{hypB}
\begin{hypB}
\label{hypB:pos_eigs}
For $u=1,2$, there are infinitely many strictly positive eigenvalues $\{\la_p(\S_u)\}_{p \geq 1}$
of $\S_u$.
\end{hypB}

The statistical analysis conducted in Section \ref{sec:results} shall demonstrate,
in the case $\ga_n \to 0$, the need to respectively recenter and rescale
(a standard statistical transformation known as \emph{studentization}) the maximum Fisher discriminant ratio, in order to get a theoretically well-grounded test statistic. These roles,
recentering and rescaling, will be played respectively by $d_1(\S_W, \ga)$
and $d_2(\S_W, \ga)$, where for a given
compact operator $\S$ with decreasing eigenvalues $\la_p$, the quantity
$d_r(\S, \ga)$ is defined for all $q \geq 1$ as
\begin{equation}
\label{eq:definition-c-d}
d_r(\S, \ga)
\eqdef
\left\{
\sump (\la_p + \ga)^{-r}  \la_p^{r}
\right\}^{1/r}
\eqsp.
\end{equation}

\subsection{Computation of the test statistics}
In practice the test statistics may be computed thanks to the kernel trick,
adapted to the kernel Fisher discriminant analysis as outlined in 
\citep[Chapter 6]{Shawe:Cristianini:2004}. Let us consider the two samples $\{ \x^{(1)}_1,\dots, \x^{(1)}_{n_1} \}$
and $\{ \x^{(2)}_{n_1},\dots, \x^{(2)}_{n_2} \}$, with $n_1 +n_2 =n$.
Denote by $\Gb^{(u)}_n: \Rset^{n_u} \mapsto \RKHS{}$, $u=1,2$, the linear operators which associates to a vector
$\al^{(u)}= [ \alpha^{(u)}_{1}, \dots, \alpha^{(u)}_{n_u}]^T$ the vector in $\RKHS{}$ given by $\Gb^{(u)}_n \al^{(u)} = \sum_{j=1}^{n_u} \alpha^{(u)}_j k(\x^{(u)}_j,\cdot)$.  This operator may be presented in a matrix form
\begin{equation}
\label{eq:definition-G}
\Gb^{(u)}_n = \left[ k(\x_1^{(u)},\cdot),\dots, k(\x_{n_u}^{(u)},\cdot) \right] \eqsp.
\end{equation}
We denote by $\Gb_n = \left[ \Gb^{(1)}_n \ \Gb^{(2)}_n \right]$.
We denote by $\Kb_n^{(u,v)} = [\Gb_n^{(u)}]^{T} \Gb^{(v)}_n$, $u,v \in \{0,1\}$,
the Gram matrix given by $\Kb_n^{(u,v)}(i,j) \eqdef k(\x^{(u)}_i,\x^{(v)}_j)$ for $i \in \{1,\dots,n_u\}$, $j \in \{1,\dots,n_v\}$.
Define, for any integer $\ell$,
$\Pb_{\ell}= \Id_{\ell}-\ell^{-1}\textbf{1}_{\ell}\textbf{1}_{\ell}^{T}$
where $\textbf{1}_{\ell}$ is the $(\ell \times 1)$ vector whose components are all equal to one and $\Id_{\ell}$ is the $(\ell \times\ell)$ identity matrix and let $\Nb_n$ be given by
\begin{equation}
\Nb_n \eqdef
\left(%
\begin{array}{cc}
  \Pb_{n_1} & 0 \\
  0 &    \Pb_{n_2} \\
\end{array}
\right).
\end{equation}
Finally, define the vector $\mb_n=(\mb_{n,i})_{1 \leq i \leq n}$ with
$\mb_{n,i}=-n_1^{-1}$ for $i=1,\dots,n_1$ and $\mb_{n,i}= n_2^{-1}$ for
$i=n_1+1,\dots,n_1+n_2$. With the notations introduced above,
$$ \hmu_2- \hmu_1 = \Gb_n\mb_n \eqsp,
\quad  \hS_u=  n_u^{-1} \Gb_n^{(u)} \Pb_{n_u} \Pb_{n_u}^{T} (\Gb_n^{(u)})^{T} \eqsp, u=1,2
\eqsp, \quad \hS_W = n^{-1} \Gb_n \Nb_n \Nb_n^{T} \Gb_n^{T} \eqsp,
$$
which implies that
\[
\pscal{\hmu_2 - \hmu_1}{(\hS_W+\gamma I)^{-1} (\hmu_2 -\hmu_1)}_{\RKHS{}}=   \mb_n^{T} \Gb_n^{T}
(n^{-1} \Gb_n \Nb_n  \Nb_n^{T} \Gb_n^{T} + \ga  \Id)^{-1} \Gb_n \mb_n \eqsp.
\]
Then, using the matrix inversion lemma, we get
\begin{align*}
&\mb_n^{T} \Gb_n^{T} (n^{-1} \Gb_n \Nb_n \Nb_n^{T} \Gb_n^{T} + \ga \Id)^{-1} \Gb_n \mb_n \\
&\quad = \ga^{-1} \mb_n^{T} \Gb_n^{T}
\left\{ \Id - n^{-1} \Gb_n \Nb_n (\gamma \Id + n^{-1} \Nb_n^{T} \Gb_n^{T}\Gb_n\Nb_n )^{-1} \Nb_n^{T} \Gb_n^{T} \right\} \Gb_n \mb_n \\
& \quad = \ga^{-1} \left\{ \mb_n^{T} \Kb_n \mb_n
 - n^{-1} \mb_n^{T}\Kb_n \Nb_n (\gamma \Id +  n^{-1} \Nb_n \Kb_n  \Nb_n )^{-1} \Nb_n  \Kb_n \mb_n \right\} \eqsp .
\end{align*}
Hence, the maximum kernel Fisher discriminant ratio may be computed from
\begin{multline*}
n_1 n_2/n \; \norm{(\hS_{W} +\gamma_n \Id)^{-1/2} (\hmu_2 - \hmu_1) }_{\H}^2 \\
=  n_1 n_2/\ga n \; \left\{ \mb_n^{T} \Kb_n \mb_n
 - n^{-1} \mb_n^{T}\Kb_n \Nb_n (\gamma \Id +  n^{-1} \Nb_n \Kb_n  \Nb_n )^{-1} \Nb_n  \Kb_n \mb_n \right\} \eqsp .
 \end{multline*}

\section{Main results}
\label{sec:results}

This discussion yields the following normalized test statistics:
\begin{equation}
\label{eq:test-statistics}
\hT_n(\ga_n) = \frac{n_1 n_2/n \; \norm{( \hS_{W} + \gamma_n \Id)^{-1/2} \hat{\delta}}^2_{\H} -d_1(\hat{\S}_W, \ga_n)}{\sqrt{2} \; d_2(\hat{\S}_W, \ga_n)} \eqsp.
\end{equation}

In this paper, we first consider the asymptotic behavior of $\hT_n$ under the null hypothesis, and
against a fixed alternative. This will establish that our nonparametric test procedure is consistent. However, this is not enough, as it can be arbitrarily slow. We thus then consider local alternatives.

For all our results, we consider two situations regarding the regularization parameter $\ga_n$; (a) a situation where $\ga_n$ is held fixed, and in which the limiting distribution is somewhat similar to the maximum mean discrepancy test statistics, and (b) a situation where $\ga_n$ tends to zero slowly enough, and in which we obtain qualitatively different results.

\subsection{Limiting distribution under null hypothesis}
Throughout this paper, we assume that the proportions $n_1/n$ and $n_2/n$ converge to strictly positive numbers,
that is
\begin{equation*}
n_u/n \to \rho_u, \quad \text{as} \; n=n_1 +n_2 \to \infty \eqsp,
\quad \text{with} \; \rho_u>0 \; \text{for} \; u=1,2 \eqsp .
\end{equation*}
In this section, we derive the distribution of the test statistics under the null hypothesis $\HH: \P_1= \P_2$ of homogeneity, which implies
$\mu_1= \mu_2$ and $\S_1 = \S_2 = \S_W$. We first consider the case where the regularization factor
is held constant $\ga_n \equiv \ga$. We denote $\dlim$ the convergence in distribution.
\begin{theo}
\label{theo:null_limit_dist_gaf}
Assume (A\ref{hypA:bounded_kernel}-B\ref{hypB:sq_sum_eigs}). Assume in addition that the probability distributions $\P_1$ and $\P_2$ are equal, \ie\ $\P_1 = \P_2 = \P$, and that $\ga_n \equiv \ga >0$. Then,
%,then for all $n \geq 1$,
\begin{equation}
\label{eq:definition-limiting-distribution}
\hT_n(\ga) \dlim  T_{\infty}(\S_W,\ga) \eqdef 2^{-1/2}\thinspace d_2^{-1}(\S_W, \ga)
\sum_{p=1}^{\infty} (\la_p(\S_W) + \ga)^{-1}\la_p(\S_W) (Z_p^2 -1) \eqsp ,
\end{equation}
where $\{\la_p(\S_W)\}_{p \geq 1}$ are the eigenvalues
of the covariance operator $\S_W$,
and $d_{2}(\S_W, \ga)$ is defined in (\ref{eq:definition-c-d}),
and  $Z_p$, $  p\geq 1 $ are independent standard normal variables.
\end{theo}

If the number of non-vanishing eigenvalues is equal to $p$ and if $\ga = 0$,
then the limiting distribution coincides with the
 limiting distribution of the Hotelling $T^2$ for comparisons of two $p$-dimensional vectors (which is
a central chi-square with $p$ degrees of freedom
~\citep{Lehmann:Romano:2005}.
The previous result is similar to what is obtained by~\citet{Gretton:Borgwardt:Rasch:Schoelkopf:Smola:2006} for the Maximum Mean Discrepancy test statistics (\MMD), we obtain a weighted sum of chi-squared distributions with \emph{summable} weights.
For a given level $\alpha \in [0,1]$, denote by $t_{1-\alpha}(\S_W,\ga)$ the $(1-\alpha)$-quantile of the distribution of $T_\infty(\S_W,\ga)$.  Then,
the sequence of test $\hT_n(\ga) \geq t_{1-\alpha}(\S_W,\ga)$, is pointwise asymptotically level $\alpha$ to test homogeneity.
Because in practice the covariance $\S_W$ is unknown, it is not possible to compute the quantile
$t_{1-\alpha}(\S_W,\ga)$. Nevertheless, this quantile can still be consistently estimated by $t_{1-\alpha}(\hS_W,\ga)$, which can be obtained from the sample covariance matrix
(see Proposition~\ref{prop:approx_limit_dist_null_gaf}).
\begin{cor}
\label{cor:null_limit_dist_gaf}
The test $\hT_n(\ga) \geq t_{1-\alpha}(\hS_W,\ga)$ is pointwise asymptotically level $\alpha$.
\end{cor}
In practice, the quantile $t_{1-\alpha}(\hS_W,\ga)$ can be  numerically computed by inverse Laplace transform \citep[see][]{Strawderman:2004,Hughett:1998}.

For all $\ga >0$, the weights $\{(\la_p + \ga)^{-1} \la_p\}_{p \geq 1}$ are summable.
However, if Assumption~(B\ref{hypB:pos_eigs}) is satisfied,
both $d_{1,n}(\ga, \S_W)$ and $d_{1,n}(\ga, \S_W)$ tend to infinity
when $n \to 0$. The following theorem shows that if $\gamma_n$ tends to zero slowly enough, 
then our test statistics is asymptotically normal:
\begin{theo}
\label{theo:null_limit_dist_gan}
Assume (A\ref{hypA:bounded_kernel}),  (B\ref{hypB:sq_sum_eigs}-B\ref{hypB:pos_eigs}).
Assume in addition that the probability distributions $\P_1$ and $\P_2$ are equal, \ie\ $\P_1= \P_2 = \P$ and that the sequence $\{\ga_n\}$
is such that
\begin{equation*}
\ga_n + d_2^{-1}(\S_W,\ga_n) d_1(\S_W,\ga_n) \ga_n^{-1}n^{-1/2} \to 0 \eqsp,
\end{equation*}
then
\begin{equation*}
\hT_n(\ga_n) \dlim \mathcal{N}(0,1) \eqsp .
\end{equation*}
\end{theo}
The proof of the theorem is postponed to Section~\ref{sec:proofs}.
Under the assumptions of Theorem~\ref{theo:null_limit_dist_gan}, the sequence of tests that rejects the null hypothesis when $\hat{T}_n(\gamma_n) \geq z_{1-\alpha}$, where $z_{1-\alpha}$ is the
$(1-\alpha)$-quantile of the standard normal distribution, is asymptotically level $\alpha$.

Contrary to the case where $\ga_n \equiv \ga$, the limiting distribution does not depend on the reproducing kernel,
nor on the sequence of regularization parameters $\{\ga\}_{n \geq 1}$. However, notice that 
$d_2^{-1}(\S_W,\ga_n) d_1(\S_W,\ga_n) \ga_n^{-1}n^{-1/2} \to 0$~requires
that $\{\ga\}_{n \geq 1}$ goes to zero at a slower rate than $n^{-1/2}$. For instance, 
if the eigenvalues $\{\la_p\}_{p \geq 1}$ decrease at a polynomial rate, 
that is if there exists $s >0$ such that we have $\la_p = p^{-s}$ for all $p \geq 1$,
then, by Lemma~\ref{lem:d_equiv}, we have $d_1(\S_W, \ga_n) 
\sim \ga_n^{-1/s}$ and $d_2(\S_W, \ga_n) \sim \ga_n^{-1/2s}$ as $n \to \infty$. 
Therefore, the condition $d_2^{-1}(\S_W,\ga_n) d_1(\S_W,\ga_n) \ga_n^{-1}n^{-1/2} \to 0$~entails in this particular case 
that $\ga_n^{-1}=o(n^{2s/1+4s})$, where the rate of decay $s$ of the eigenvalues of the covariance operator $\S_W$,
depends \emph{both on the kernel and the underlying distribution $\P_1= \P_2 = \P$}.   
Besides, it may seem surprising that the limiting distribution is normal.
This is due to two facts. First, we regularize the sample covariance operator prior to inversion
(being of finite rank, the inverse of $\hS$ is obviously not defined). Second, the problem is here truly infinite dimensional, because we have assumed that the eigenvalues are infinite dimensional
$\lambda_p(\S_W) > 0$ for all $p$ \citep[see][Theorem 14.4.2, for a related result]{Lehmann:Romano:2005}.

\subsection{Limiting behavior against fixed alternatives}
We study the power of the test based on $\widehat{T}_n(\gamma_n)$ under alternative hypotheses. The minimal requirement is  to prove that this sequence of tests is consistent.
A sequence of tests of constant level $\alpha$  is said to be \emph{consistent in power} if the probability of accepting the
null hypothesis of homogeneity goes to zero as the sample size goes to infinity under a \emph{fixed} alternative.
Recall that two probability  $\P_1$ and $\P_2$ defined on a measurable space $(\Xset,\Xsigma)$ are called \emph{singular} if there exist two disjoint sets $A$ and $B$ in $\Xsigma$
whose union is $\Xset$ such that $\P_1$ is zero on all measurable subsets of $B$ while $\P_2$ is zero on all measurable subsets of $B$. This is denoted by $\P_1 \perp \P_2$.

When \condgaf~or when $\ga_n \to 0$, and $\P_1$ and $\P_2$ are not singular,
then the following proposition shows that the limits in both cases are finite,
strictly positive and independent of the kernel otherwise
\citep[see][for similar results for canonical correlation analysis]{Fukumizu:Gretton:Sunn:Schoelkopf:2008}.
The following result gives some useful insights on $\Vert \S_W^{-1/2} (\mu_2 - \mu_1) \Vert_{\H}$,
the population counterpart of $\Vert(\hS_W +\ga_n \Id)^{-1/2}(\hmu_2 - \hmu_1)\Vert_{\H}$ on which
our test statistics is based upon.

\begin{prop}
\label{prop:imagekernel}
Assume (A\ref{hypA:bounded_kernel}-A\ref{hypA:Kernel:injection}).
Let $\nu$ a measure dominating $\P_1$ and $\P_2$, and let
$p_1$ and $p_2$ the densities of
$\P_1$ and $\P_2$ \wrt\ $\nu$.
The norm $\Vert \S_W^{-1/2} (\mu_2 - \mu_1) \Vert_{\H}$ is infinite
if and only if $\P_1$ and $\P_2$ are mutually singular.
If $\P_1$ and $\P_2$ are nonsingular, $\Vert \S_W^{-1/2} (\mu_2 - \mu_1) \Vert_{\H}$ is finite
and is given by
$$
\norm{ \S_W^{-1/2} (\mu_2 - \mu_1) }_{\H}^2 =
\frac{1}{\rho_1 \rho_2} \left( 1 - \int \frac{p_1 p_2}{\rho_1 p_1 + \rho_2 p_2} d\nu \right)
\left( \int \frac{p_1 p_2}{ \rho_1 p_1 + \rho_2 p_2 } d\nu \right) ^{-1}.
$$
It is equal to zero if the $\chi^2$-divergence is null, that is, if and only if $\P_1=\P_2$.
\end{prop}

By combining the two previous propositions, we therefore obtain the following consistency theorem:

\begin{theo}
\label{theo:test_power_consistency}
 Assume (A\ref{hypA:bounded_kernel}-A\ref{hypA:Kernel:injection}).
 Let $\P_1$ and $\P_2$ be two distributions over $(\Xset,\Xsigma)$,
 such that $\P_2 \ne \P_1$. If either  \condgaf~or \condgan, then
 for any $t >0$
\begin{equation}
\label{eq:power_consist_gaf}
\P_{\hh}(\hT_n(\ga) > t) \to 1 \eqsp .
\end{equation}
\end{theo}

\subsection{Limiting distribution against local alternatives}
When the alternative is fixed, any sensible test procedure will have  a power that tends to one as the sample size $n$
tends to infinity. This property is not suitable for comparing the limiting power of different test procedures. Several approaches are possible to answer this question.
One such approach  is to consider sequences of \emph{local alternatives}~\citep{Lehmann:Romano:2005}. Such alternatives tend to the null
hypothesis as $n \to \infty$ at a rate which is such that the limiting distribution of sequence the test statistics under the sequence of alternatives
converge to a non-degenerate random variable. To compare two sequences of tests for a given sequence of alternatives, one may then compute the ratio of the
limiting powers, and choose the test which has the largest power.

In our setting, let $\P_1$ denote a fixed probability on $(\Xset,\Xsigma)$ and let $\P_2^n$
be a sequence of probability on $(\Xset,\Xsigma)$. The sequence $\P_2^n$ depends on the sample size $n$ and
converge to $\P_1$ as $n$ goes to infinity with respect to a certain distance. In the asymptotic analysis of our test statistics against sequences of local alternatives,
the $\chi^2$-divergence $\chisqdiv{\P_1}{\P_2^n}$ is defined for all $n$ as
\begin{equation}
\label{eq:definition-eta-n}
\eta_n \eqdef \norm{\frac{d \P_2^n}{d \P_1} - 1}_{\ltwo(\P_1)} \eqsp,
\end{equation}
for $\P_2^n$ absolutely continuous \wrt\ $\P_1$. Therefore, in the subsequent sections,
we shall make the following assumption:
\begin{hypC}
\label{hypC}
For any $n$, $\P_2^n$ is absolutely continuous \wrt\ $\P_1$, 
and $\chisqdiv{\P_1}{\P_2^n} \to 0$ as $n$ tends to infinity.
\end{hypC}
The following theorem shows that under local alternatives, we get a series of shift in the chi-squared distributions when \condgaf:
\begin{theo}
\label{theo:alt_limit_dist_gaf}
 Assume (A\ref{hypA:bounded_kernel}), (B\ref{hypB:sq_sum_eigs}), and (C).
 Assume in addition  $\ga_n \equiv \ga >0$ and that $n \eta_n^2 = O(1)$,
%  and if there exists $C>0$ such that $\lim_{n \to \infty} \sum_{p =1}^{\infty} (\la_p + \ga)^{-1} \dep^2 = C$
 then
\begin{equation*}
 \widehat{T}_n(\gamma) \dlim
\thinspace 2^{-1/2}d_2^{-1}(\S_1, \gamma)
 \sum_{p=1}^{\infty} (\la_p(\S_1)+\gamma)^{-1}\la_p(\S_1) \{(Z_p + a_{n,p}(\ga))^2 -1 \} \eqsp ,
\end{equation*}
with
\begin{equation}
\label{eq:def_anp}
a_{n,p}(\ga)=  (n_1 n_2/n)^{1/2}
\pscal{(\S_1 +\ga \Id)^{-1/2}(\mu_2^n -\mu_1)}{e
_p}_\mathcal{H}\eqsp ,
\end{equation}
 where $\{Z_p \}_{p \geq 1}$ are independent standard normal random variables,
 defined on a common probability space.
 \end{theo}

When the sequence of regularization parameters $\{\ga_n\}_{n \geq 1}$ tends to zero
at a slower rate than $n^{-1/2}$, the test statistics is shown to be asymptotically normal,
with the same limiting variance as the one under the null hypothesis, but with 
a non-zero limiting mean,
as detailed in the next two results. While the former states the asymptotic normality
under general conditions, the latter highlights  the fact that the asymptotic mean-shift
in the limiting distribution may be conveniently expressed from the limiting $\chi^2$-divergence
of $\P_1$ and $\P_2^n$ under additional smoothness assumptions on the spectrum
of the covariance operator.

 \begin{theo}
\label{theo:alt_limit_dist_gan}
Assume (A\ref{hypA:bounded_kernel}),
and (B\ref{hypB:sq_sum_eigs}-\ref{hypB:pos_eigs}), and (C).
Let $\{\gamma_n\}_{n \geq 1}$ be a sequence such that 
\begin{align}
\label{eq:condition-gamma-dist-gan}
&\condgan \\
&d_2^{-1}(\S_1,\ga_n) n \eta_n^2 = O(1) \quad \text{and} \quad d_2^{-1}(\S_1,\ga_n) d_1(\S_1,\ga_n) \eta_n \to 0 \eqsp,
\end{align}
where $\{\eta_n\}_{n \geq 1}$ is defined in~(\ref{eq:definition-eta-n}). If the following limit exists,
 \begin{equation}
 \label{eq:expasymp}
 \meanshift \eqdef \lim_{n \to \infty} \frac{n  \norm{(\S_1 + \gamma_n \id )^{-1/2} (\mu_2^n - \mu_1)}^2_{\H}} {d_2(\S_1,\ga_n)}
 \eqsp,
 \end{equation}
then,
 \begin{equation*}
\widehat{T}_n(\ga_n) \dlim \gauss(\rho_1 \rho_2 \meanshift , 1) \eqsp .
 \end{equation*}
\end{theo}
 
\begin{cor}
Under the assumptions of Theorem \ref{theo:alt_limit_dist_gan}, if there exists $a >0$ such that
\begin{equation*}
\pscal{ \mu_2^n - \mu_1}{\S_1^{-1-a} (\mu_2^n - \mu_1)}_{\H} < \infty \eqsp,
\end{equation*}
and if the following limit exists,
  \begin{equation*}
 \meanshift  =\lim_{n \to \infty} d_2(\S_1,\gamma_n)^{-1} n  \eta_n^2  \eqsp ,
 \end{equation*}
then,
 \begin{equation*}
\widehat{T}_n(\ga_n) \dlim \gauss(\rho_1 \rho_2 \meanshift , 1) \eqsp .
 \end{equation*}
\end{cor}
 
It is worthwhile to note that $\rho_1 \rho_2 \Delta$, the limiting mean-shift
of our test statistics against sequences of local alternatives does not depend on the choice 
of the reproducing kernel. This means that, at least in the large-sample setting $n \to \infty$,
the choice of the kernel is irrelevant, provided that for some $a>0$ we have $\pscal{ \mu_2^n - \mu_1}{\S_1^{-1-a} (\mu_2^n - \mu_1)}_{\H} < \infty$. Then, we get that the sequences of local alternatives converge to the null at rate $\eta_n = C\: d_2^{1/2}(\S_1,\ga_n) n^{-1/2}$ for some constant $C>0$, which is slower than the usual parametric rate $n^{-1/2}$ since $ d_2(\S_1,\ga_n) \to \infty$ as $n \to \infty$ as shown in Lemma~\ref{lem:c-d-go-to-infinity}. Note also that conditions of the form $\pscal{\mu_2^n -\mu_1}{\S_1^{-1-\alpha}(\mu_2^n -\mu_1)}_{\H} < \infty$ imply that the sequence of local alternatives are limited to \emph{smooth enough} densities $p_2^n$ around $p_1$.

\section{Discussion}
\label{sec:discussion}

We illustrate now the behaviour of the limiting power of our test statistics
against two different types of sequences of local alternatives. Then, we compare 
the power of our test statistics against the power of the Maximum Mean Discrepancy test statistics proposed by~\cite{Gretton:Borgwardt:Rasch:Schoelkopf:Smola:2006}. Finally, 
we highlights some links between testing for homogeneity and supervised binary classification.

\subsection{Limiting power against local alternatives of KFDA}

We have seen that our test statistics is consistent in power against fixed alternatives, for both regularization schemes $\ga_n \equiv \ga$ and $\ga_n \to 0$. We shall now examine the behaviour of the power of our test statistics, against different types of sequences of local alternatives: i) directional alternatives, ii) non-directional alternatives. For this purpose, we consider a specific reproducing kernel, the periodic spline kernel, whose derivation is given below. Indeed, when $\P_1$ is the uniform distribution on $[0, 1]$, and $d\P_2 /d\P_1= 1 + \eta c_q$ with $c_q$ is a one-component contamination on the Fourier basis, we may conveniently compute 
a closed-form equivalent when $n \to \infty$ of the  eigenvalues of the covariance operator $\S_1$, and therefore the power function of the test statistics. 

\paragraph{Periodic spline kernel}
The periodic spline kernel, described in~\citep[Chapter 2]{Wahba:1990},
is defined as follows. Any function $f$ in $\ltwo(\Xset)$, where $\Xset$ 
is taken as the torus $\mathbb{R}/2 \pi \mathbb{Z}$, may be expressed
in the form of a Fourier series expansion
$f(t) = \sum_{p=0}^{\infty} a_p c_p(t)$
where $\sum_{p=0}^{\infty} a_p^2$, and for all $\ell \geq 1$
\begin{align}
\label{eq:cp_def}
c_0(t) &=\mathbf{1}_{\Xset} \\
c_{2\ell -1}(t) &=\sqrt{2}\;\sin(2 \pi (2\ell -1) t) \\
c_{2\ell}(t) &=\sqrt{2}\; \cos(2 \pi (2\ell -1) t) \eqsp .
\end{align}
Let us consider the family of RKHS defined by 
$\H^m = \{ f:\; f \in \ltwo(\Xset),\; \sum_{p=0}^{\infty} \la_p^{-1} a_p^2 < \infty \}$
 with $m>1$, 
where $\la_p = (2 \pi p)^{-2m}$ for all $p \geq 1$, 
whose norm is defined for all $f \in \ltwo(\Xset)$ as
\begin{equation}
\norm{f}_{\H}^2 = 1/2 \; \sum_{p=0}^{\infty} (2\pi p)^{-2m} a_p^2 \eqsp .
\end{equation}
Therefore, the associated reproducing kernel $k(x,y)$ writes as 
\begin{equation*}
k_{m}(x,y) = 2 \sum_{p=0}^{\infty} (2 \pi p)^{-2m} c_p(x-y) 
			 = \frac{(-1)^{m-1}}{(2m)!} B_{2m} \left( (x-y) - \lfloor x-y \rfloor \right) \eqsp ,
\end{equation*}
where $B_{2m}$ is the $2m$-th Bernoulli polynomial.

The set $\{ e_p(t), p \geq 1 \}$ is actually an orthonormal basis of $\H$,
where $e_p(t) \eqdef \la_p^{1/2} c_p(t)$ for all $p \geq 1$.
Let us consider $\P_1$ the uniform probability measure on $[0,1]$.
We have $e_p - \E_{\P_1}[e_p] \equiv e_p$ and $\mu_1 \equiv 0$,
where $\mu_1$ is the mean element associated with $\P_1$. Hence,
$\{ (\la_p, e_p(t)), p \geq 1 \}$ is an eigenbasis of $\S_1$ the covariance 
operator associated with $\P_1$, where for all $\ell \geq 1$
\begin{align}
\label{eq:lap_def}
\la_0  &=1 \\
\la_{2\ell -1} &=(4 \pi \ell)^{-2m} \\
\la_{2\ell} &=(4 \pi \ell)^{-2m} \eqsp .
\end{align}

Note that the parameter $m$ characterizes the RKHS $\H^{m}$ and its associated reproducing kernel $k_{m}(\cdot,\cdot)$, 
and therefore controls the rate of decay of the eigenvalues of the covariance operator $\S_1$. 
Indeed, by Lemma~\ref{lem:d_equiv}, we have $d_1(\S_1, \ga_n) = C_1\:\ga_n^{-1/2m}$ and 
$d_2(\S_1, \ga_n) = C_2\:\ga_n^{-1/4m}$ for some constants $C_1, C_2 >0$ as $n \to \infty$.

\paragraph{Directional alternatives}
Let us consider the limiting power of our test statistics in the following setting:
\begin{equation}
\label{eq:pitman_alt}
\HH:\; \P_1= \P_2^n \quad \text{against}
\quad \hhn:\; \P_1 \neq \P_2^n ,\; \; \text{with}\; \P_2^n \; 
\text{such that} \; d\P_2^n /d\P_1 = 1 + A n^{-1/2}c_q \eqsp,
\end{equation}
where $\P_1$ is the uniform probability measure on $[0,1]$,
and $c_q(t)$ is defined in~(\ref{eq:cp_def}). In the case $\ga_n \equiv \ga$, 
given a significance level $\alpha \in (0,1)$, the associated critical level $t_{1-\alpha}$ is defined as
satisfying 
\begin{equation*}
\P\left(2^{-1/2}d_2^{-1}(\S_1,\ga)
 \sum_{p=1}^{\infty} (\la_p(\S_1)+\ga)^{-1}\la_p(\S_1) \{Z_p^2 -1 \} 
 > t_{1-\alpha}\right) =\alpha \eqsp .
\end{equation*}
Note that $a_{n,p}(\ga)=0$ for all $p \geq 1$ (from Theorem~\ref{theo:alt_limit_dist_gaf}) except for $p=q$ where 
\begin{equation*}
a_{n,q}(\ga)=\sqrt{A} \; \sqrt{n_1 n_2/n^2} \thinspace (\la_q + \ga)^{-1/2} \la_q^{1/2}\eqsp .
\end{equation*} 

\begin{figure}[ht]
  \hfill
  \begin{minipage}[t]{.32\textwidth}
    \includegraphics[scale=0.48]{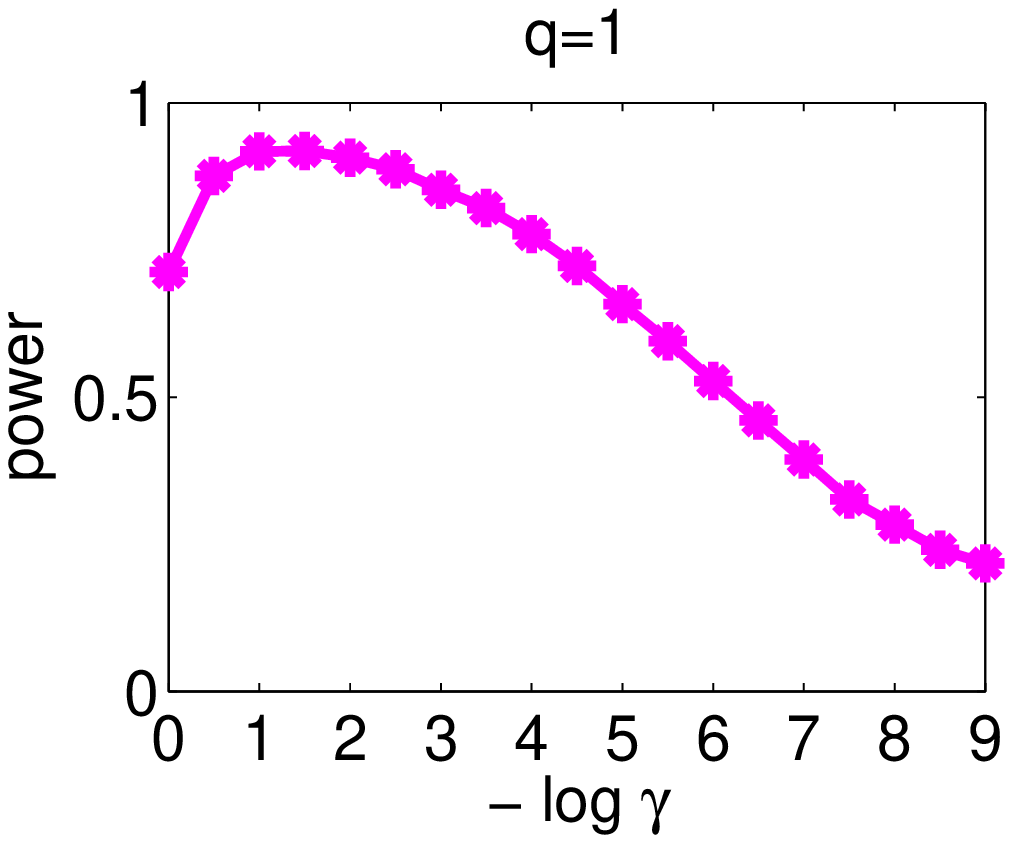}
  \end{minipage}
  \hfill
  \begin{minipage}[t]{.32\textwidth}
    \includegraphics[scale=0.48]{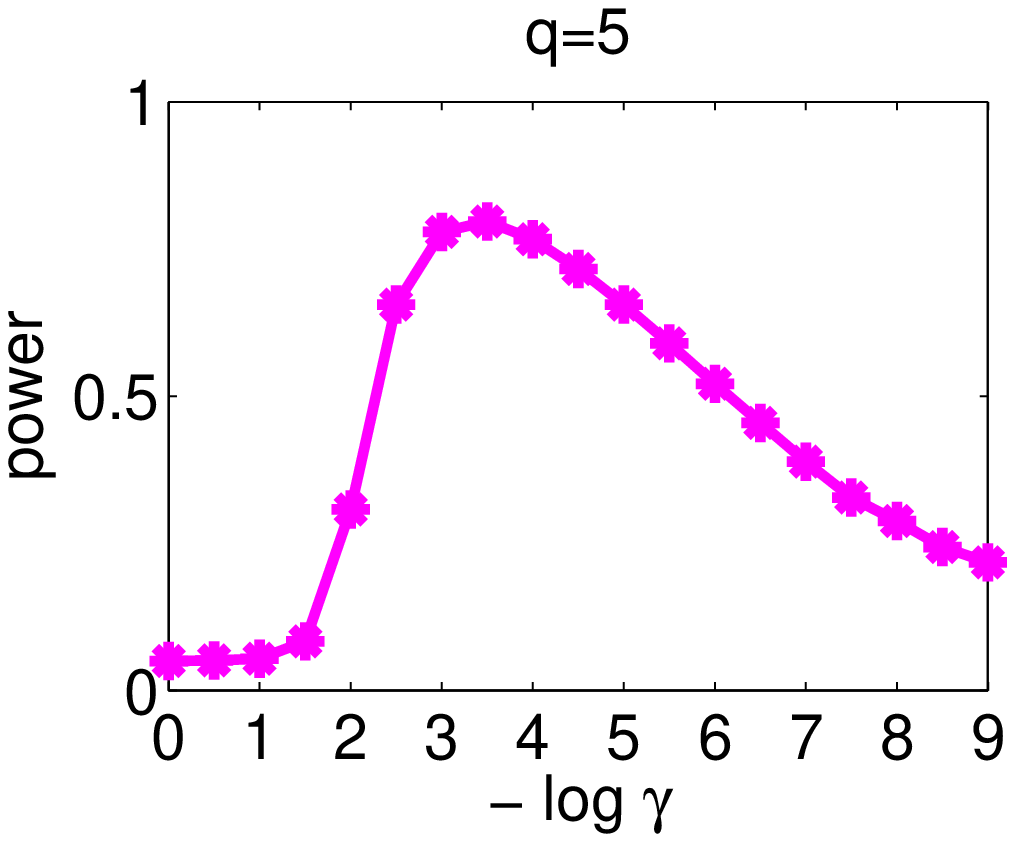}
  \end{minipage}
  \hfill
  \begin{minipage}[t]{.32\textwidth}
    \includegraphics[scale=0.48]{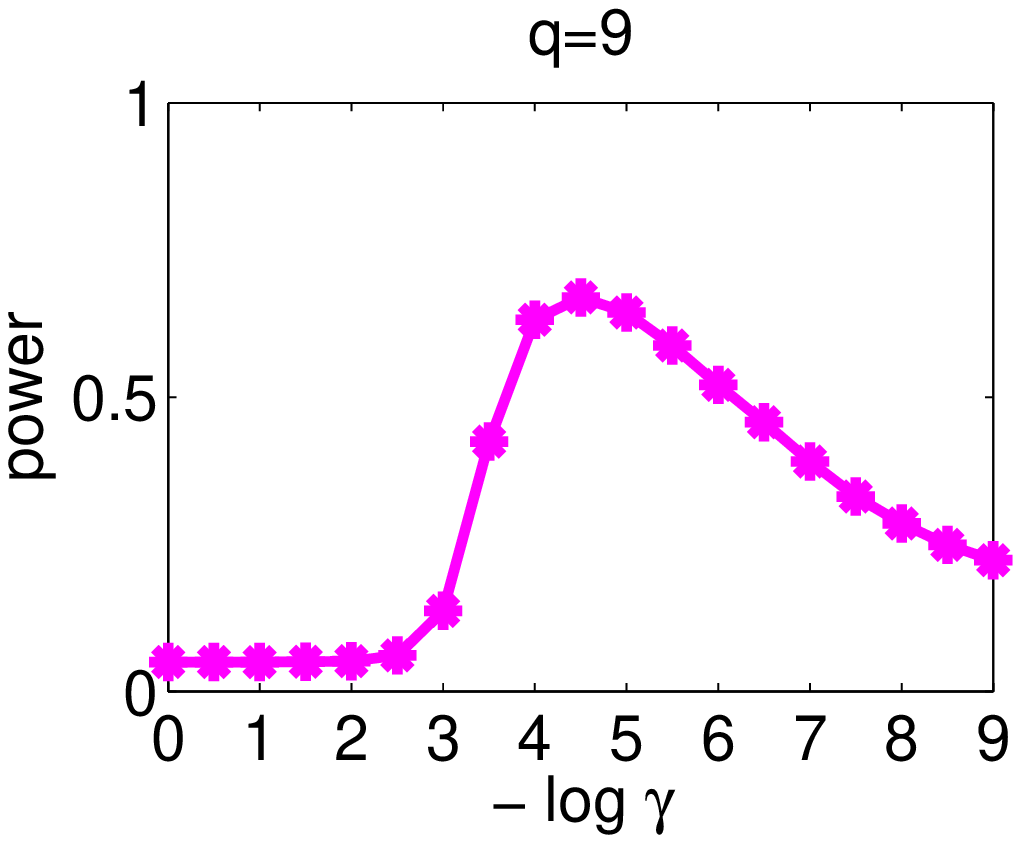}
  \end{minipage}
  \hfill
   \vspace*{.2cm}
    \caption{Evolution of power of \KFDA~as $\ga =1,10^{-1},\dots,10^{-9}$,
    for $q$-th component alternatives with (from left to right) with $q=1,5,9$.}
    \label{fig:power_gaf_kfda_qs}
\end{figure}

In order to analyze the behaviour of the power for varying values of $\ga$
and for different values of $q$, we compute the limiting power, when taking $m=2$
in the periodic reproducing kernel, and for $q=1,5,9$, and investigate the evolution
of the power as a function of the regularization parameter $\ga$. As Figure~\ref{fig:power_gaf_kfda_qs} shows,
our test statistics has trivial power, that is equal to $\alpha$, when $\ga \gg \la_q$,
while it reaches stricly nontrivial power as long as $\ga \leq \la_q$. 
This motivates the study of the decaying regularization scheme $\ga_n \to 0$ of our test statistics, in order to incorporate the $\ga \to 0$ into our large-sample framework. 
In the next paragraph, we shall demonstrate that the version of our test statistics with decaying regularization parameter $\ga_n \to 0$ reaches high power against a 
broader class of local alternatives, which we call \emph{non-directional alternatives}, 
where $q \equiv q_n \to \infty$, as opposed to \emph{directional alternatives} where $q$
was kept constant. Yet, for having nontrivial power with the test statistics $\hT(\ga_n)$ against such sequences of local alternatives, the non-directional sequence of local alternatives have to converge to the null at a slower rate than $\sqrt{n}$.

\paragraph{Non-directional alternatives}
Now, we consider the limiting power of our test statistics in the following setting:
\begin{equation}
\label{eq:shifting_freq_alt}
\HH:\; \P_1= \P_2^n \quad \text{against}
\quad \hhn:\; \P_1 \neq \P_2^n ,\; \; \text{with}\; \P_2^n \; 
\text{such that} \; d\P_2^n /d\P_1 = 1 + \eta_n c_{q_n} \eqsp,
\end{equation}
Assume $\P_1$ is the uniform probability measure on $[0,1]$,
and consider again the periodic spline kernel of order $2m$.
Take $\{q_n \}_{n \geq 1}$ a nonnegative nondecreasing sequence of integers.
Now, if the sequence of local alternatives is converging to the null
at rate $\eta_n = (2\Delta)^{1/2} q_n^{1/4} n^{-1/2}$ for some $\Delta >0$,
with $q_n =o(n^{1/1+4m})$ for our asymptotic analysis to hold,
then as long as $\ga_n \equiv \la_{q_n} = q_n^{-2m}$ we have
\begin{align*}
\lim_{n \to \infty} \; \P_{\hhn}\left(\hT_n(\ga_n) > z_{1-\alpha}\right) 
&= \P_{\hhn}\left( Z_1 + \rho_1 \rho_2 \Delta > z_{1-\alpha} \right) \\
&= 1- \Phi\left[z_{1-\alpha} -\rho_1 \rho_2 \Delta \right]  \eqsp .
\end{align*}
where we used Lemma~\ref{lem:d_equiv} together with Theorem~\ref{theo:alt_limit_dist_gan}.
On the other hand, if $\ga_n^{-1} q_n^{-2m} =o(1)$, then the limiting power 
is trivial and equal to $\alpha$. 

Back to the fixed-regularization test statistics $\hT_n(\ga)$,
we may also compute the limiting power of $\hT_n(\ga)$
against the non-directional sequence of local alternatives defined in~(\ref{eq:shifting_freq_alt})
by taking into account Remark~\ref{remark:trunc_neg_relax} to use Theorem~\ref{theo:alt_limit_dist_gaf}.
Indeed, as $n$ tends to infinity, since 
$a_{n,q_n}(\ga) = (\rho_1 \rho_2)^{1/2} (\la_{q_n} + \ga)^{-1/2} \la_{q_n} \eta_n$, then the fixed-regularization
version $\hT_n(\ga)$ of the test statistics has trivial power against non-directional alternatives. 

\begin{remark}
We analyzed the limiting power of our test statistics in the specific case
where $\P_1$ is the uniform distribution on $[0,1]$ and the reproducing kernel belongs to 
the family of periodic spline kernels. Yet, our findings carry over more general settings 
as illustrated by Table~\ref{table:decays}. Indeed, for general distributions with polynomial decay in the tail and (nonperiodic) gaussian kernels, the eigenvalues of the covariance operator
still exhibit similar behaviour as in the example treated above. 
\end{remark}

\begin{table}
  \centering
  \begin{tabular}{|c|c|c|c|}
  \hline 
  
                      & $\la_p(\S)$           &  $d_{1}(\S,\ga)$ & $d_{2}(\S,\ga)$\\ 
                      \hline \hline
    Normal tails    & $O\left(\exp (- c p^{1/d})\right)$ &  $O\left(\log^d( 1/\ga)\right)$ & $O\left(\log^{d/2}( 1/\ga)\right)$ \\
    \hline 
    Polynomial tails  & $O\left(p^{-\beta} \right)$ for any $\beta > \alpha$ & $O\left(\ga^{-1/\beta}\right)$
    & $O\left(\ga^{-1/2 \beta}\right)$\\ 
    \hline
     \end{tabular}
  \caption{examples of rate of convergence for the gaussian kernels for $\Xset= \Rset^p$  }
  \label{table:decays}
  \end{table}

We now discuss the links between our procedure with the previously proposed Maximum Mean Discrepancy (MMD) test statistics. We also highlight interesting links with supervised kernel-based classification.

\subsection{Comparison with Maximum Mean Discrepancy}
Our test statistics share many similarities with the Maximum Mean Discrepancy test statistics
of~\cite{Gretton:Borgwardt:Rasch:Schoelkopf:Smola:2006}. In the case \condgaf~,
both have limiting null distribution which may be expressed as an infinite weighted mixture
of chi-squared random variables. Yet, while $\hT_n^{\text{MMD}} \dlim C \sump \la_p (Z_p^2 -1)$
where $\hT_n^{\text{MMD}}$ denotes the test statistics used by MMD,
we have in our case
$\hT_n^{\text{\KFDA}}(\ga_n) \dlim C \sump (\la_p + \ga_n)^{-1}\la_p (Z_p^2 -1)$. Roughly speaking,
the test statistics based on \KFDA~uniformly weights the components associated with the first eigenvalues
of the covariance operator, and downweights the remaining ones, which allows to gain greater
power for testing by focusing on the user-tunable number of components of the covariance operator.
On the other hand, the test statistics based on MMD is naturally sensitive to differences lying
on the first components, and gets progressively less sensitive to differences in higher components.
Thus, our test statistics based on \KFDA~allows to give equal weights to differences lying
in (almost) all components, the effective number of components on the which the test statistics focus on
being tuned \textit{via} the regularization parameter $\ga_n$. These differences may be illustrated by considering the behavuour of MMD against sequences of local alternatives respectively with fixed-frequency and non-directional, for periodic kernels.

\paragraph{Directional alternatives}
Let us consider the setting defined in~(\ref{eq:pitman_alt}). By a similar reasoning, we may
also compute the limiting power of $\hT_n^{\text{MMD}}$ against directional
sequences of local alternatives, with a periodic spline kernel of order $m=2$,
for different components $q=1,5,9$. Both test statistics \KFDA~and \MMD~ reach high power
when the sequences of local alternatives lies on the first component. However, the power of MMD tumbles down for higher-order alternatives whereas the power of \KFDA~remains strictly nontrivial for high-order alternatives as long as $\ga$ is sufficiently small.

\begin{figure}[ht]
  \hfill
  \begin{minipage}[t]{.32\textwidth}
    \includegraphics[scale=.48]{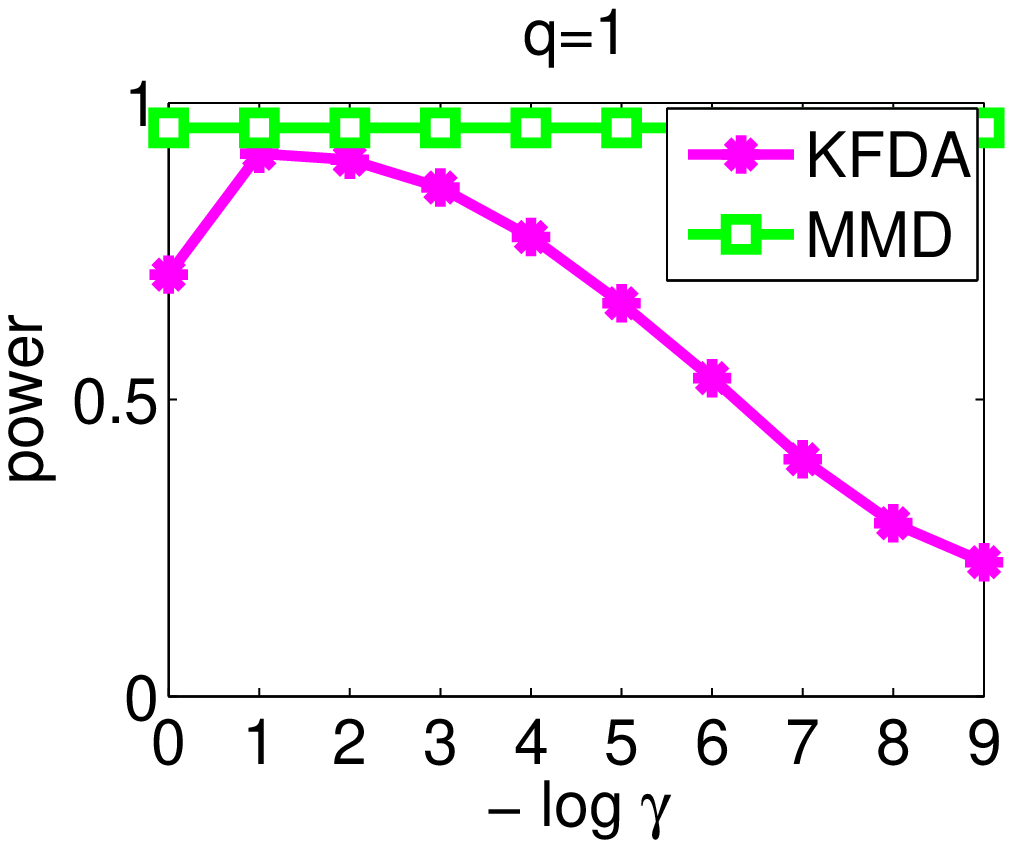}
  \end{minipage}
  \hfill
  \begin{minipage}[t]{.32\textwidth}
    \includegraphics[scale=.48]{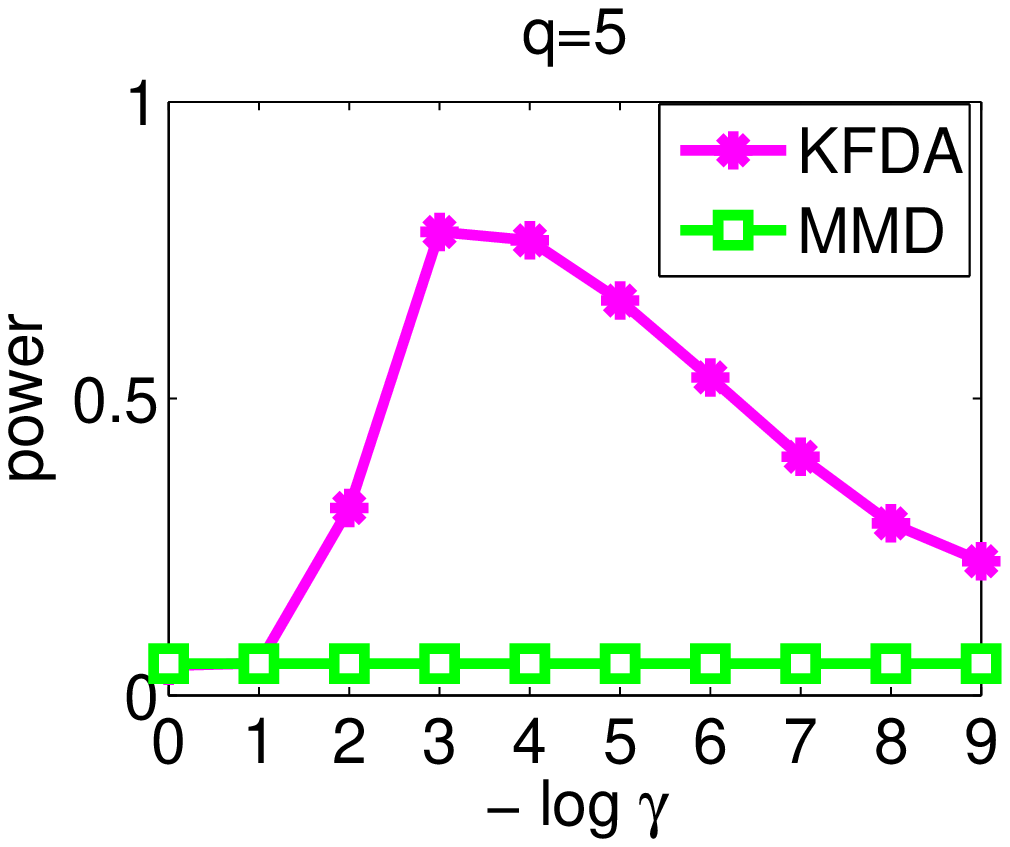}
  \end{minipage}
  \hfill
  \begin{minipage}[t]{.32\textwidth}
    \includegraphics[scale=.48]{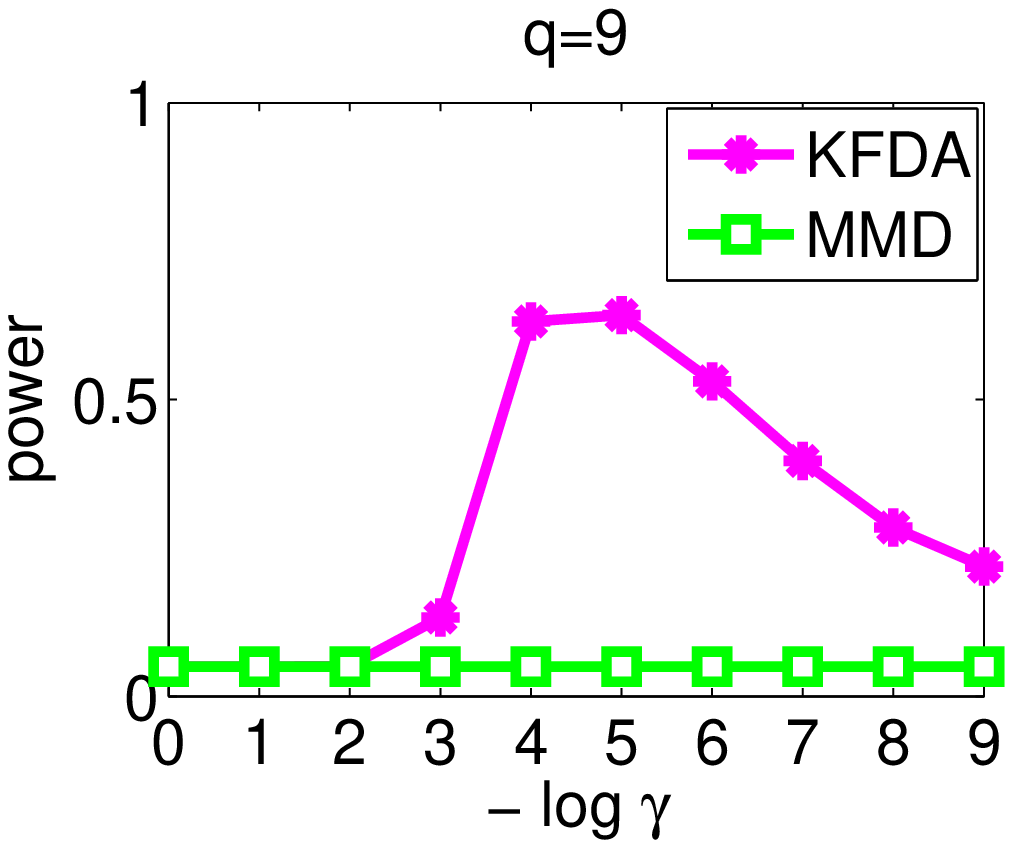}
  \end{minipage}
  \hfill
   \vspace*{.2cm}
    \caption{Comparison of the evolution of power of \KFDA~versus the power of \MMD~
    as $\ga =1,10^{-1},\dots,10^{-9}$,
    for $q$-th component alternatives with (from left to right) with $q=1,5,9$.}
\end{figure}

\paragraph{Non-directional alternatives}
Now, consider sequences of local alternatives as defined in~(\ref{eq:shifting_freq_alt}). 
The test statistics MMD does not notice such alternatives. Therefore, MMD has trivial power equal to $\alpha$ against non-directional alternatives. 

\subsection{Links with supervised classification}

When the sample sizes of each sample are equal, that is when $n_1 = n_2$,
\KFDA~is known to be equivalent to Kernel Ridge Regression (KRR), 
also referred to as smoothing spline regression in statistics.
In this case, KRR performs a kernel-based least-square regression fit on the labels, where
the samples are respectively labelled $-1$ and $+1$. The recentering parameter $d_1(\S_1,\ga_n)$ in our procedure coincides with the so-called \emph{degrees of freedom} in smoothing spline regression, which were often advocated to provide a relevant measure of complexity
for model selection~\citep[see][]{Efron:2004}. In particular, since the mean-shift in the limiting normal distribution against local alternatives is lower-bounded by
$n d_{1}^{-1}(\S_1,\ga_n)  \langle  (\mu_2 - \mu_1), (\S_1 + \gamma_n \id )^{-1} (\mu_2 - \mu_1) \rangle$, this suggests an algorithm for selecting
$\ga_n$ and the kernel. For a fixed degree of freedom $d_{1}(\S_1,\ga_n)$, maximizing
the asymptotic mean-shift (which corresponds to the class separation) is likely to yield greater power. As future work,
we plan to investigate, both theoretically and practically, the use of (single and multiple) kernel learning procedures as developed by~\cite{Bach04}
for maximizing the expected power of our test statistics in specific applications.

\section{Experiments}\label{sec:experiments}

In this section, we investigate the experimental performances of our test statistic \KFDA,
and compare it in terms of power against other nonparametric test statistics.

\subsection{Speaker verification}

We conducted experiments in a speaker verification task~\cite{Reynolds:2004}, on a subset of 8 female speakers using data from the NIST 2004 Speaker Recognition Evaluation.
We refer the reader to~\citep{Louradour:Daoudi:Bach:2007} for instance for details on the pre-processing of data. The figure shows
averaged results over all couples of speakers. For each couple of speaker, at each run we took $3000$ samples of each speaker and launched our \KFDA-test
to decide whether samples come from the same speaker or not, and computed the type II error by comparing the prediction to ground truth. We averaged the results for $100$ runs for each couple, and all couples of speaker. The level was set to $\alpha= 0.05$,
and the critical values were computed by a bootstrap resampling procedure.
Since the observations may be considered dependent within the sequences,
and independent between the sequences, we used a fixed-block variant of the boostrap,
which consists in using boostrap samples built by piecing together several boostrap samples drawn in each sequence.  We performed the same experiments for the
 Maximum Mean Discrepancy and the Tajvidi-Hall test statistic (TH). We summed up the results
by plotting the ROC-curve for all competing methods. Our method reaches good empirical power for a small value of the prescribed level
($1-\beta=90\%$ for $\alpha=0.05\%$). Maximum Mean Discrepancy also yields good empirical performance on this task.
\begin{figure}
    \centering
    \includegraphics[scale=.5]{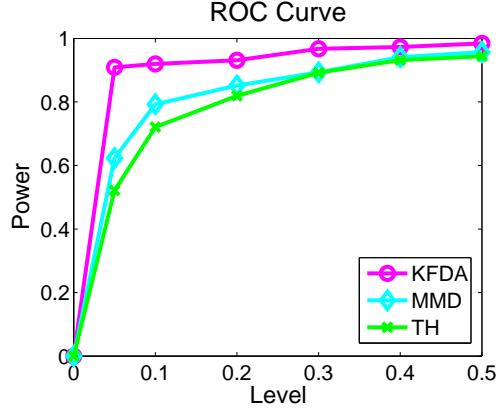}
    \caption{Comparison of ROC curves in a speaker verification task}
    \label{fig:roc_speaker}
\end{figure}

\section{Conclusion}
We proposed a well-calibrated kernel-based test statistic for testing the homogeneity 
of two samples, built on the kernel Fisher discriminant analysis algorithm, for which we proved that the asymptotic limit distribution under null hypothesis is standard normal distribution when de regularization parameter decays to zero at a slower rate than $n^{-1/2}$. Besides, our test statistic can be readily computed from Gram matrices once a reproducing kernel is defined,
and reaches nontrivial power aqgainst a large class of alternatives under mild conditions on the regularization parameter. Finally, our \KFDA-test statistic yields competitive performance for speaker identification purposes.

%%%%%%%%%%%%%%%%%%%%%%%%%%%%%%%%%%%%%%%%%%%%%%%%%%%%%%%%%%%%%%%%%%%%%
%%%%%%%%%%%%%%%%%%%%%%%%%%%%%%%%%%%%%%%%%%%%%%%%%%%%%%%%%%%%%%%%%%%%%
%%%%%%%%%%%%%%%%%%%%%%%%%%%%%%%%%%%%%%%%%%%%%%%%%%%%%%%%%%%%%%%%%%%%%

\section{Proof of some preliminary results}
\label{sec:prelim}
We preface the proof by some useful results relating the \KFDA~statistics to kernel independent quantities.
\begin{prop}
\label{prop:range}
Assume (A\ref{hypA:bounded_kernel})-(A\ref{hypA:Kernel:injection}).
Let $\P_1$ and $\P_2$ be two probability distributions on $(\Xset,\Xsigma)$,
and denote by $\mu_1, \mu_2$ the associated mean (see \eqref{eq:definition-moyenne}).
Let $\Q$ be a probability dominating $\P_1$ and $\P_2$, and let $\S$ be the associated covariance operator
(see \eqref{eq:definition-covariance}). Then,

$$
\left\| \frac{d \P_1}{d \Q} -  \frac{d \P_2}{d \Q} \right\|_{\ltwo(\Q)}  < \infty,
$$
if and only if the vector $(\mu_2 - \mu_1) \in  \H$ belongs to the range of the square root $\S^{1/2}$. In addition,
\begin{equation}
\label{eq:equivalence-norme}
 \pscal{\mu_2 - \mu_1}{\S^{-1} (  \mu_2 - \mu_1  )}_{\H}  =    \left\| \frac{d \P_1}{d \Q} -  \frac{d \P_2}{d \Q} \right\|^2_{\ltwo(\Q)}   \eqsp.
\end{equation}

\end{prop}
\begin{proof}
Denote by $\{\lambda_k\}_{k \geq 1}$ and $\{e_k\}_{k \geq 1}$ the strictly positive eigenvalues and the corresponding eigenvectors of the covariance operator $\S$,
respectively.    For $k \geq 1$, set
\begin{equation}
\label{eq:definition-fk}
f_k= \lambda_k^{-1/2} \left\{ e_k - \Q e_k \right\} \eqsp.
\end{equation}
By construction, for any $k,\ell \geq 1$,
\[
\lambda_k \delta_{k,\ell}
= \pscal{e_k}{\Sigma e_\ell}_{\H} = \pscal{e_k - \Q e_k}{e_\ell - \Q e_\ell}_{\ltwo(\Q)} =  \lambda_k^{1/2} \lambda_\ell^{1/2}   \pscal{f_k}{f_\ell}_{\ltwo(\Q)} \eqsp,
\]
where $\delta_{k,\ell}$ is Kronecker's delta. Hence $\{f_k\}_{k \geq 1}$ is an orthonormal system of $\ltwo(\Q)$.
Note  that $\mu_2 - \mu_1$ belongs to the range of $\S^{1/2}$ if and only if
\begin{enumerate}[(a)]
\item \label{item:orthogonality} $\pscal{\mu_2 - \mu_1}{g}_{\H}=0$ for all $g$ in the null space of $\Sigma$,
\item \label{item:rangecondition} $ \pscal{\mu_1-\mu_2}{\Sigma^{-1}(\mu_1-\mu_2)}_{\H} \eqdef \sum_{p=1}^\infty \lambda_p^{-1} \pscal{e_p}{(\mu_1-\mu_2)}_{\H}^{2} < \infty$.
\end{enumerate}
Consider first condition \eqref{item:orthogonality}. For any $g \in \H$, il follows from the definitions that
\begin{multline*}
\pscal{\mu_2-\mu_1}{g}_{\H{}}  = \int \left( d \P_1 - d \P_2 \right) g = \int \left( d \P_1 - d \P_2 \right) ( g - \Q g) \\
= \pscal{\frac{d \P_1}{d \Q}- \frac{d \P_2}{d \Q}}{g - \Q g}_{\ltwo(\Q)} \eqsp.
\end{multline*}
If $g$ belongs to the null space of $\Sigma$, then $\| g - \Q g \|_{\ltwo(\Q)}= 0$, and the previous relation implies that
$\pscal{\mu_2-\mu_1}{g}_{\H{}}= 0$. Consider now \eqref{item:rangecondition}.
\begin{multline}
\label{eq:maha_leq_chi2div}
\sum_{p=1}^\infty \lambda_p^{-1} \pscal{e_p}{(\mu_1-\mu_2)}_{\H}^2
= \sum_{p=1}^\infty \lambda_p^{-1} \left( \int \{d \P_1(x) - d \P_2(x) \} e_p(x)  \right)^2 \\
 =  \sum_{p=1}^\infty \pscal{\frac{d \P_1}{d \Q} - \frac{d \P_2}{d \Q}}{f_p}_{\ltwo(\Q)}^2 \leq
 \left\| \frac{d \P_1}{d \Q} - \frac{d \P_2}{d \Q} \right\|_{\ltwo(\Q)}^2 \eqsp.
\end{multline}

In order to prove the equality, we simply notice that because of the density of the RKHS in $L^2(\Q)$, then   $\{f_k \}_{k \geq 1}$ is a complete orthonormal basis of the space of functions  $\ltwo_0(\Q)$, defined as
\begin{equation}
\label{eq:definition-L_0^2(Q)}
\ltwo_0(\Q) \eqdef \left\{ g \in \ltwo(\Q) \eqsp, \int (g - \Q g)^2 d \Q > 0 \quad \text{and} \quad \Q g = 0 \right\} \eqsp.
\end{equation}
\end{proof}

\begin{lem}
\label{lem:covariance-contingency}
Assume (A\ref{hypA:bounded_kernel})-(A\ref{hypA:Kernel:injection}).
Let $\P_1$ and $\P_2$ two probability distributions on $(\Xset,\Xsigma)$ such that $\P_2 \ll \P_1$.

Denote by $\Sigma_1$ and $\Sigma_2$ the associated covariance operators. Then, for any $\gamma > 0$,
\begin{align}
\label{eq:hs_chi2div_ineq}
\hsnorm{\Id -\S_1^{-1/2} \S_2^n \S_1^{-1/2}}^2 &\leq 4  \norm{\frac{d \P_2}{d \P_1} - 1}_{\ltwo(\P_1)}^2 \eqsp , \\
\label{eq:trace_to_mscont}
\left|\trace{\{(\S_1 + \gamma \Id )^{-1}(\S_2 - \S_1)\}}\right|
&\leq 2 d_2(\S_1,\gamma) \norm{\frac{d \P_2}{d \P_1} - 1}_{\ltwo(\P_1)}
\eqsp .
\end{align}
where $d_2(\S_1,\gamma)$ is defined in \eqref{eq:definition-c-d}.
\end{lem}

\begin{proof}
Denote by $\{\lambda_k\}_{k \geq 1}$ and $\{e_k\}_{k \geq 1}$ the strictly positive eigenvalues and the corresponding eigenvectors of the covariance operator $\S_1$.  Note that $\pscal{e_k}{\S_1 e_{\ell}}=\lambda_k \delta_{k , \ell}$ for all $k$ and $\ell$.
Let us denote $f_k= \lambda_k^{-1/2} \left\{e_k - \P_1 e_k \right\}$. Then,
we have $\pscal{f_k}{f_{\ell}}_{\ltwo(\P_1)} = \delta_{k , \ell}$. Note that
\begin{align*}
&\quad \sum_{k,\ell=1}^{\infty}  \left\{ \delta_{k,\ell}
- \la_k^{-1/2}\la_{\ell}^{-1/2} \pscal{e_k}{\S_2e_{\ell}}_{\H} \right\}^2
 \\
&= \sum_{k,\ell=1}^{\infty}  \left\{
   \pscal{f_k}{\left(1-\frac{d \P_2}{d \P_1}\right) f_{\ell}}_{\ltwo{(\P_1)}}
 + \la_k^{-1/2}\la_{\ell}^{-1/2}
 \pscal{\mu_2 -\mu_1}{e_k}_{\H}
 \pscal{\mu_2 -\mu_1}{e_{\ell}}_{\H}
 \right\}^2 \eqsp .\\
\end{align*}
Then, using that $(a+b)^2 \leq 2(a^2+b^2)$, and (\ref{eq:maha_leq_chi2div})
in Proposition \ref{prop:range} with $\S = \S_1$, we obtain
\begin{equation}
\label{eq:hs_chi2div_ineq_inside}
\hsnorm{\Id -\S_1^{-1/2} \S_2^n \S_1^{-1/2}}^2 \leq 4  \norm{\frac{d \P_2}{d \P_1} - 1}_{\ltwo(\P_1)}^2 \eqsp .
\end{equation}
Denote, for all $p,q \geq 1$
\begin{equation}
\label{eq:epq_def}
\epq \eqdef \pscal{e_p}{(\S_1^{-1/2}\S_2\S_1^{-1/2} - \Id)e_q} \eqsp .
\end{equation}
By applying the H\"{o}lder inequality, and using (\ref{eq:hs_chi2div_ineq}), we get
\begin{multline*}
\left|\trace{\{(\S_1 + \gamma \Id )^{-1}(\S_2 - \S_1)\}}\right|
= \sump \left|  \pscal{e_p}{(\S_1 + \ga\Id )^{-1}\S_1 e_p} \epp \right| \\
\leq \left( \sump  \pscal{e_p}{(\S_1 + \ga \Id )^{-1}\S_1 e_p}^2 \right)^{1/2} \left(\sump   \epp^2 \right)^{1/2}
\leq 2 d_2(\S_1,\ga) \norm{\frac{d \P_2}{d \P_1} - 1}_{\ltwo(\P_1)}  \eqsp , \\
\end{multline*}
which completes the proof of (\ref{eq:trace_to_mscont}).
\end{proof}

\begin{prop}
\label{prop: trace_conv_cov_op}
Assume (A\ref{hypA:bounded_kernel}). Let $\{X_1^{n}, \dots, X_n^{n}\}$ be a triangular array of i.i.d random variables,
whose mean element and covariance operator are respectively $(\mu^{n}, \S^{n})$.
If, for all $n$ all the eigenvalues $\lambda_p(\S^{n})$ of $\S^{n}$ are non-negative,
and if there exists $C >0$ such that for all $n$ we have $\sum_{p=1}^\infty \lambda^{1/2}_p(\S^{n}) < C$,
then $\sum_{p=1}^{\infty} |\la_p(\hat{\S} -\S^{n})|=O_{P}(n^{-1/2})$.
\end{prop}

\begin{proof}
Lemma~\ref{lem:borne-sum-vp} shows that, for  any orthonormal basis  $\{e_p\}_{p \geq 1}$ in the RKHS $\H$:
\[\sum_{p=1}^\infty |\lambda_p(\hS-\S^n)| \leq \sum_{p=1}^\infty \norm{(\hS-\S^n)e_p}_{\H} \eqsp.
\]
We take the orthonormal family of eigenvectors
$\{e_p\}_{p \geq 1}$ of the covariance operator $\S^n$ (associated to the eigenvalues $\lambda_p(\S^n)$ ranked in decreasing order).
Then, it suffices to show that $\sum_{p=1}^\infty \norm{(\hS-\S^n)e_p}_{\H} = O_{P}(n^{-1/2})$.
Note that,
\[
\left( \hS - \S^n \right) e_p = n^{-1} \sum_{i=1}^{n} \zeta_{p,n,i} - \left(n^{-1} \sum_{i=1}^{n} k(X_i,\cdot)\right) \; \left(n^{-1} \sum_{i=1}^{n} \ce_{p,n}(X_i)\right) \eqsp,
\]
where $\ce_{p,n}= e_p - \E^n [ e_p(X_1) ]$ and
\[
\zeta_{p,n,i} \eqdef  k(X_i,\cdot) \ce_{p,n}(X_i) - \E^n \left\{ k(X_1,\cdot) \ce_{p,n}(X_1) \right\}
\]
By the Minkowski inequality,
 \begin{multline*}
 \left\{ \E^n \norm{\left( \hS - \S^n \right) e_p }_{\H}^2 \right\}^{1/2} \leq
 \left\{ \E^n \norm{ n^{-1} \sum_{i=1}^{n} \zeta_{p,n,i}^2}  \right\}^{1/2} \\
 +
 \left\{ \E^n \left[ \norm{n^{-1} \sum_{i=1}^{n} k(X_i,\cdot) }_{\H}^2 \; \left|n^{-1} \sum_{i=1}^{n}
 \ce_{p,n}(X_i)\right|^2 \right] \right\}^{1/2} = A_1+A_2 \eqsp.
 \end{multline*}
We consider these two terms separately. Consider first $A_1$. We have
\[
A_1^2 = n^{-1} \E^n \norm{\zeta_{p,n,i}}^2_{\H} \leq n^{-1} \E^n \left\{ \norm{k(X_1,\cdot)}^2_{\H} |\ce_{p,n}
(X_1)|^2 \right\}
\leq n^{-1} |k|_\infty \E^n \left[ |\ce_{p,n}(X_1)|^2 \right] \eqsp.
\]
 Consider now $A_2$. Since
$\norm{n^{-1} \sum_{i=1}^{n} k(X_i,\cdot)}^2_{\H} \leq |k|_\infty$, we have
\[
A_2^2 \leq n^{-1} |k|_\infty \E^n \left[ |\ce_{p,n}(X_1)|^2 \right] \eqsp.
\]
This shows, using the Minkowski inequality, that
\[
\left\{ \E^n \left( \sum_{p=1}^\infty \norm{(\hS - \S^n)e_p}_{\H} \right)^2\right\}^{1/2} \leq 2 |k|_\infty^{1/2} n^{-1/2} \sum_{p=1}^\infty \left\{ \E^n \left[ |\ce_{p,n}(X_1)|^2 \right]  \right\}^{1/2} \eqsp.
\]
Since by assumption
$  \sum_{p=1}^\infty \left\{ \E^n \left[ |\ce_{p,n}(X_1)|^2 \right]  \right\}^{1/2}
= \sum_{p=1}^\infty \lambda^{1/2}_p(\S^{n}) < \infty$,
the proof is concluded.
\end{proof}

\begin{cor}
\label{cor:hs_conv_cov_op}
Assume (A\ref{hypA:bounded_kernel}). Let $\{X_{1,n_1}^{(1)}, \dots, X_{n_1,n_1}^{(1)}\}$
and $\{X_{1,n_2}^{(2)}, \dots, X_{n_2,n_2}^{(2)}\}$ be two triangular arrays,
whose mean elements and covariance operators are respectively $(\mu_{1}^{n}, \S_{1}^{n})$
and $(\mu_{2}^{n}, \S_{2}^{n})$, where $n_1/n \to \rho_1$ and $n_2/n \to \rho_2$ as
$n$ tends to infinity.  If  $\sup_{n \geqslant 0} \sum_{p=1}^\infty \lambda^{1/2}_p(\S_a^{n}) < \infty$,
then
\begin{equation}
\label{eq:tr_conv_cov_op_eq}
\sum_{p=1}^\infty |\lambda_p(\hS_W -\S_W)| =O_{P}(n^{-1/2}) \eqsp .
\end{equation}
In addition, we also have
\begin{equation}
\label{eq:hs_conv_cov_op_eq}
\hsnorm{\hS_W -\S_W} =O_{P}(n^{-1/2}) \eqsp .
\end{equation}
\end{cor}

\begin{proof}
Since $\hat{\S}_{W} -\S_{W} = n_1 n^{-1} (\hat{\S}_{1} -\S_{1}) + n_2 n^{-1} (\hat{\S}_{2} -\S_{2}^n)$, then
\begin{equation*}
\sum_{p=1}^\infty \norm{(\hS_W-\S_W)e_p}_{\H} \leq n_1 n^{-1} \sum_{p=1}^\infty \norm{(\hS_1-\S_1)e_p}_{\H}
+ n_2 n^{-1} \sum_{p=1}^\infty \norm{(\hS_2-\S_2^n)e_p}_{\H} \eqsp ,
\end{equation*}
and applying twice Proposition \ref{prop: trace_conv_cov_op} leads to \eqref{eq:tr_conv_cov_op_eq}.
Now, using that
\begin{equation}
\hsnorm{\hS_W -\S_W} \leq \sum_{p=1}^\infty |\lambda_p(\hS_W -\S_W)| \eqsp ,
\end{equation}
then \eqref{eq:hs_conv_cov_op_eq} follows as a direct consequence of \eqref{eq:tr_conv_cov_op_eq}.
\end{proof}

\section{Asymptotic approximation of the test statistics}
\label{sec:proofs}
The following proposition shows that in the asymptotic study of our test statistics, we can replace most empirical quantities by population quantities. For ease of notation, we shall denote
$\mu_2 -\mu_1$ by $\delta$.
$\hmu_2 -\hmu_1$ by $\hat{\delta}$.
\begin{prop}
\label{prop:approximation-statistique}
Assume (C). If
\begin{align*}
&\condgan \\
&d_2^{-1}(\S_1,\ga_n) n \eta_n^2 = O(1) \quad \text{and} \quad d_2^{-1}(\S_1,\ga_n) d_1(\S_1,\ga_n) \eta_n \to 0 \eqsp,
\end{align*}
then, $\hT_n(\ga_n)= \tT_n(\ga_n) + o_P(1)$, where
\begin{equation}
\label{eq:definition-tT_n}
\tT_n(\ga)
\eqdef  \frac{(n_1 n_2/n)\;
\norm{ \left( \S_1 + \ga \Id \right)^{-1/2} \hat{\delta} }_{\H}^2
 -d_{1}(\S_1, \ga)}{\sqrt{2} d_{2}(\S_1, \ga)} \eqsp.
\end{equation}
\end{prop}

\begin{proof}
Notice that
  \begin{align*}
 |d_{2}(\hS_W, \ga_n) - d_{2}(\S_1, \ga_n)| &\leq |d_{2}(\hS_W, \ga_n) - d_{2}(\S_W, \ga_n)| + |d_{2}(\S_W, \ga_n) - d_{2}(\S_1, \ga_n)|
\eqsp .
\end{align*}
Then, on the one hand, using Eq.~\eqref{eq:trace_perturb_carleman} for $r=2$
in Lemma~\ref{lem:perturb_cov} with $S = \S_W $ and $\Delta = \hS_W - \S_W$ and Eq.~\eqref{eq:tr_conv_cov_op_eq} in Corollary \ref{cor:hs_conv_cov_op},
we get $\left| d_{2}(\hS_W, \ga_n) - d_{2}(\S_W, \ga_n) \right| = O_P(\ga_n^{-1} n^{-1/2})$. On the other hand,
using Eq.~\eqref{eq:trace_perturb_hs_d2} in Lemma~\ref{lem:perturb_cov} with $S = \S_1$
and $\Delta = n_2 n^{-1} (\S_2^n - \S_1)$, we get
$d_{2}(\S_W, \ga_n) - d_{2}(\S_1, \ga_n) =  O(\eta_n)$.
Furthermore, similar reasoning, using Eq.~\eqref{eq:trace_perturb_carleman} and Eq.~\eqref{eq:trace_perturb_hs_d1}
again in Lemma~\ref{lem:perturb_cov} allows to prove that $d_{2}^{-1}(\S_1, \ga_n)d_{1}(\hS_W, \ga_n)
= d_{2}^{-1}(\S_1, \ga_n)d_{1}(\S_1, \ga_n)  + o_P(1)$.
Next, we shall prove that
\begin{equation}
\label{eq:approx_leading_term}
\norm{ ( \hS_W + \gamma_n \Id )^{-1/2} \hat{\delta}}_{\H}^2
= \norm{ ( \S_1 + \gamma_n \Id  )^{-1/2} \hat{\delta}}_{\H}^2
+ n^{-1} O_{P}\left\{ (d_{1}(\S_1, \ga_n)+ n\eta_n^2)(\gamma_n^{-1} n^{-1/2} + \eta_n) \right\} \eqsp.
\end{equation}
Using straightforward algebra, we may write
\begin{equation}
\label{eq:big_decomp}
\left|
\norm{ ( \hS_W + \gamma_n \Id )^{-1/2} \hat{\delta}}_{\H}^2
- \norm{ ( \S_1 + \gamma_n \Id  )^{-1/2} \hat{\delta}}_{\H}^2
\right|
\leq  A_1 A_2 \; \{B_1 + B_2\} \eqsp ,
\end{equation}
with
\begin{eqnarray*}
A_1 &\eqdef\norm{(\S_1 + \gamma_n \Id )^{-1/2} \hdelta}_{\H} \eqsp ,
\quad \quad B_1&\eqdef \norm{ (\hS_W + \gamma_n \Id )^{-1/2}  ( \hS_W - \S_W) ( \S_1 + \gamma_n \Id  )^{-1/2} } \eqsp ,\\
A_2 &\eqdef  \norm{(\hS_W + \gamma_n \Id )^{-1/2} \hdelta}_{\H} \eqsp ,
\quad \quad B_2&\eqdef  \norm{ (\hS_W + \gamma_n \Id )^{-1/2}  ( \S_2^n - \S_1) ( \S_1 + \gamma_n \Id  )^{-1/2} }  \eqsp .\\
\end{eqnarray*}
We now prove that
\begin{align}
\label{eq:A1_eq}
A_1^2 &=  O_{P}( n^{-1}d_{1}(\S_1, \ga_n)+ \eta_n^2) \eqsp, \\
\label{eq:A2_eq}
A_2^2 &=  O_{P}( n^{-1}d_{1}(\S_1, \ga_n)+ \eta_n^2) \eqsp.
\end{align}
We first consider (\ref{eq:A1_eq}). Note that $\E \left(\hat{\delta} \otimes \hat{\delta} \right)
 = \dan \otimes \dan + n_1^{-1} \Sigma_1 + n_2^{-1} \Sigma_2^n$, which yields
 \begin{multline}
 \label{eq:zaid}
\E  \| (\S_1 + \gamma_n \Id )^{-1/2}  \hat{\delta} \|^2
 = \trace{ \left\{ (\S_1 + \gamma_n \Id )^{-1}  \E \left( \hat{\delta} \otimes \hat{\delta} \right) \right\} } =
 \pscal{\dan}{(\S_1 + \gamma_n \Id )^{-1} \dan}_{\H}
\\ + \frac{n}{n_1 n_2}\trace{ \left\{ (\S_1 + \gamma_n \Id )^{-1} \S_1 \right\} }
 + n_2^{-1} \trace{ \left\{ (\S_1 + \gamma_n \Id )^{-1} \left(\S_2^n -\S_1 \right)\right\} } \eqsp .
 \end{multline}
Using Proposition~\ref{prop:range} with $\S = \S_1$ together with Assumption (C),
we may write
  \begin{equation*}
  \left| \pscal{\dan}{(\S_1 + \gamma_n \Id )^{-1} \dan}_{\H} \right|
  \leq  \left| \pscal{\dan}{\S_1^{-1} \dan}_{\H} \right| \leq \norm{\frac{d \P_2^n}{d \P_1} - 1}_{\ltwo(\P_1)}^2 = \eta_n^2 \eqsp .
 \end{equation*}
Next, applying Lemma~\ref{lem:covariance-contingency},
we obtain
\begin{equation*}
\left|\trace{\{(\S_1 + \gamma_n \Id )^{-1}(\S_2^n - \S_1)\}}\right|
= O(d_{2}(\S_1, \ga_n)\eta_n) \eqsp ,
\end{equation*}
which yields
 \begin{equation}
 \label{eq:var_delta_sigma1}
 \E  \| (\S_1 + \gamma_n \Id )^{-1/2}  \hat{\delta} \|^2
  = (n/n_1 n_2)  d_{1}(\S_1, \ga_n)\left\{1 + O(\eta_n)\right\} + O(\eta_n^2) \eqsp .
 \end{equation}
Finally, we get (\ref{eq:A1_eq}) by the Markov inequality. Now, to prove (\ref{eq:A2_eq}), it suffices to observe that
$ \openorm{ (  \hS_W + \gamma_n \Id  )^{-1} ( \S_1 + \gamma_n \Id ) } = 1 +  o_P(1)$,
and then conclude from (\ref{eq:A1_eq}).
Next, using the upper-bound $\openorm{(  \S + \gamma_n \Id  )^{-1/2} } \leq \gamma_n^{-1/2}$, and Corollary \ref{cor:hs_conv_cov_op} which gives $\hsnorm{\hS_W - \S_W} = O_{P}(n^{-1/2})$, we get
\begin{equation}
\label{eq:B1_eq}
B_1=O_{P}(\gamma_n^{-1} n^{-1/2}) \eqsp .
\end{equation}
Finally, under Assumption (C), using Eq.~\eqref{eq:hs_chi2div_ineq}
in Lemma~\ref{lem:covariance-contingency}, we obtain
\begin{equation}
\label{eq:B2_eq}
B_2 =  O_P(\eta_n) \eqsp .
\end{equation}
The proof of (\ref{eq:approx_leading_term}) is concluded by plugging
(\ref{eq:A1_eq}-\ref{eq:A2_eq}-\ref{eq:B1_eq}-\ref{eq:B2_eq})
into (\ref{eq:big_decomp}).
\end{proof}

\begin{remark}
\label{remark:approx_assumption_relax}
For the sake of generality, we proved the approximation result under the assumptions $\condgan$ on the one hand, $d_2^{-1}(\S_1,\ga_n) n \eta_n^2 = O(1)$ and $d_2^{-1}(\S_1,\ga_n) d_1(\S_1,\ga_n) \eta_n \to 0$ on the other hand. However, in the case $\ga_n \equiv \ga$, the approximation is still valid if 
$n \eta_n^3 \to 0$, which allows to use this approximation to derive the limiting power 
of our test statistics against non-directional sequences of local alternatives as in~(\ref{eq:shifting_freq_alt}).
\end{remark}

\section{Proof of Theorems \ref{theo:alt_limit_dist_gaf}-\ref{theo:alt_limit_dist_gan}}

For ease of notation, in the subsequent proofs, we shall often omit $\S_1$ in quantities involving it.
Hence, from now on, $\la_p, \la_q, d_2$ stand for $\la_p(\S_1), \la_q, d_2(\S_1,\ga)$. Define
\begin{equation}
\label{eq:definition-Ynp}
Y_{n,p,i} \eqdef
\begin{cases}
\left( \frac{n_2}{n_1 n} \right)^{1/2} \left(e_p(X_i^{(1)}) - \E [e_p(X_1^{(1)}) ] \right)
& 1 \leq i \leq n_1 \eqsp,\\
- \left( \frac{n_1}{n_2 n} \right)^{1/2} \left(e_p(X_{i-n_1}^{(2)}) - \E [e_p(X_1^{(2)}) ] \right)
& n_1+ 1 \leq i \leq n \eqsp.
\end{cases}
\end{equation}
The following lemma
gives formulas for the moments of $Y_{n,p,i}$, used throughout the actual proof of the main results.

\begin{lem}
\label{lem:formulae}
Consider $\{ Y_{n,p,i} \}_{1 \leq i \leq n, p \geq 1}$  and
%$\{ M_{n,p,i} \}_{2 \leq i \leq n, p \geq 1}$
as defined respectively
in (\ref{eq:definition-Ynp})
%and (\ref{eq:definition-Mnp})
. Then
\begin{align}
\label{eq:formulae:sum}
\sumi \E[ Y_{n,p,i} Y_{n,q,i}] &= \lambda_p^{1/2} \lambda_q^{1/2}\{\delta_{p,q} + n_1 n^{-1} \epq\}   \\
\label{eq:borne-covariance-carre}
\PCov{(Y_{n,p,i}^2, Y_{n,q,i}^2)} &\leq C n^{-2} \kmax \la_p^{1/2} \la_q^{1/2}
   (1 + \epp )^{1/2} ( 1 + \varepsilon_{q,q})^{1/2} \eqsp .
\end{align}
\end{lem}

\begin{proof}
The first expressions are proved by elementary calculations from
\begin{align*}
&  \E[ Y_{n,p,1}  Y_{n,q,1}] =  \sqfacone \delta_{p,q} \lambda_p (\S_1) \\
&  \E[ Y_{n,p,1}  Y_{n,q,n_1+1}] = 0, \quad \quad \text{since} \quad X_1^{(1)} \perp X_1^{(2)} \\
&  \E[ Y_{n,p,n_1+1}  Y_{n,q,n_1+1}] = \sqfactwo  \la_p^{1/2} \la_q^{1/2} \left\{\delta_{p,q}+  \pscal{e_p}{(\S_1^{-1/2}\S_2^n\S_1^{-1/2} - \Id)e_q}\right\} \eqsp .
\end{align*}
Next, notice that, for all $p \geq 1$, we have by the reproducing property and the the Cauchy-Schwarz inequality
\[
|e_p(x)| = \pscal{e_p}{k(x,\cdot)}_{\H{}} \leq \norm{e_p}_{\H{}} \norm{k(x,\cdot)}_{\H{}} \leq \kmax^{1/2} \eqsp .
\]
which yields
\begin{align*}
\left| \PCov( Y_{n,p,i}^2,Y_{n,q,i}^2) \right| &\leq \E [ Y_{n,p,i}^2 Y_{n,q,i}^2 ] + \E[Y_{n,p,i}^2] \E[ Y_{n,q,i}^2]\\
 &\leq C \E^{1/2} [Y_{n,p,i}^4] \E^{1/2}[Y_{n,q,i}^4] \\
 &\leq Cn^{-1} \kmax \E^{1/2} [Y_{n,p,i}^2] \E^{1/2}[Y_{n,q,i}^2]\\
  &\leq Cn^{-2} \kmax \la_p^{1/2} \la_q^{1/2}
   (1 + \epp )^{1/2}
   ( 1 + \varepsilon_{q,q})^{1/2} \eqsp . \qedhere
\end{align*}
\end{proof}

\subsection{Proof of Theorem \ref{theo:alt_limit_dist_gaf}}
\begin{proof}
The proof is adapted from~\cite[pages 195-199]{Serfling:1980}. By Proposition \ref{prop:approximation-statistique},
\[
\hT_n(\ga) =  \frac{ \hV_{n,\infty}(\ga) - d_{1}(\S_1, \ga) }{ \sqrt{2} d_{2}(\S_1, \ga)} + o_P(1) \eqsp ,
\]
where
\begin{equation*}
\label{eq:sum_rec}
\hV_{n,\infty}(\ga) \eqdef \sum_{p=1}^\infty \left( \la_p + \ga \right)^{-1}
\left( S_{n,p} + \sqrt{ \frac{n_1 n_2}{n} } \dep \right) ^2  \eqsp ,
\end{equation*}
with
\begin{equation}
\label{eq:definition-Snp}
S_{n,p} \eqdef \sqrt{\frac{n_1 n_2}{n}} \pscal{\hdelta - \dan}{e_p} = \sum_{i=1}^{n} Y_{n,p,i} \eqsp .
\end{equation}
Now put
\begin{equation}
\hV_{n,N}(\ga) \eqdef \sum_{p=1}^N \left( \la_p + \ga \right)^{-1}
\left( S_{n,p} + \sqrt{ \frac{n_1 n_2}{n} } \dep \right) ^2  \eqsp.
\end{equation}
Because $\{Y_{n,p,i}\}$ are zero mean, independent, Lemma~\ref{lem:formulae}-Eq.~\eqref{eq:formulae:sum} shows that, as $n$ goes to infinity, $\sum_{i=1}^n \PCov(Y_{n,p,i},Y_{n,q,i}) \to \lambda_p^{1/2} \la_q^{1/2} \delta_{p,q}$ .
In addition, the Lyapunov condition is satisfied, since using \eqref{eq:borne-covariance-carre},
$\sum_{i=1}^n \E[Y_{n,p,i}^4]  \leq C n^{-1} \lambda_p$.
We may thus apply the central limit theorem for multivariate triangular arrays,
which yields $\mathbf{S}_{n,N} \dlim \mathcal{N}(0, \mathbf{\Lambda}_N)$
where $\mathbf{S}_{n,N} = (S_{n,1}, \dots, S_{n,N})$ and
$(\mathbf{\Lambda}_N)_{p,q}=\delta_{p,q} \la_p$, $1 \leq p,q \leq N$.
Fix $u$ and let $\epsilon >0$ be given.
Then, using the version of the continuous mapping theorem
stated in \cite[Theorem 18.11]{vdVaart:1998},
with the sequence of quadratic functions
$\{ g_n\}_{n \geq 1} $ defined as $[\;g_n:  \mathbf{T}_N=(T_{1}, \dots, T_N) \mapsto ( \mathbf{T}_N + \ab_n)^{T} [\text{diag}(\alpha_1, \dots, \alpha_N)] ( \mathbf{T}_N + \ab_n) \; ]$,
we may write
\begin{equation}
\label{eq:clt_trunc}
| \E[ \rme^{ \rmi u \hV_{n,N}(\ga)} ] -\E[ \rme^{\rmi u V_{n,N}(\gamma)} ] | \leq \epsilon \eqsp ,
\end{equation}
with $V_{n,N}(\gamma) \eqdef \sum_{p=1}^{N} (\la_p+\ga)^{-1}\la_p (Z_p + a_{n,p})^2$,
where $\{Z_p \}_{p \geq 1}$ are independent standard normal random variables,
defined on a common probability space, and $\{a_{n,p}\}_{p \geq 1}$ are defined in (\ref{eq:def_anp}).
Next, we prove that $ \lim_{N \to \infty} \limsup_{n \to \infty} \E[(\hV_{n,\infty}(\ga) -\hV_{n,N}(\ga))^2] = 0$. By the Rosenthal inequality (see \cite[theorem 2.12]{Petrov:1995},
there exists a constant $C$ such that $\E[ S_{n,p}^4 ] \leq C (n^{-1} \lambda_p + \lambda_p^2)$.  The Minkowski inequality then leads to
\begin{align*}
\label{eq:rem_hat_trunc}
&\quad \E^{1/2}[(\hV_{n,\infty}(\ga) -\hV_{n,N}(\ga))^2]\\
&\leq  \thinspace \sum_{p=N+1}^{\infty} (\la_p+\ga)^{-1}
\thinspace \E^{1/2}\left\{\left( S_{n,p} + \sqrt{ \frac{n_1 n_2}{n} } \dep \right)^4\right\} \\
&\leq  C \left\{\ga^{-1} \sum_{p=N+1}^{\infty} \la^{1/2}_p (n^{-1/2} + \la^{1/2}_p)
+ n \sum_{p=N+1}^{\infty} (\la_p+\ga)^{-1}  \dep^2 \right\} \\
&\leq  C \left\{\ga^{-1} \sum_{p=N+1}^{\infty} \la_p^{1/2}
+ n \sum_{p=N+1}^{\infty} (\la_p+\ga)^{-1}  \dep^2 \right\} + o(1)\eqsp .\\
\end{align*}
Notice that, using (\ref{eq:maha_leq_chi2div})
in Proposition \ref{prop:range} with $\S = \S_1$, we have
\begin{equation}
\label{eq:trunc_neg}
n \sum_{p=N+1}^\infty (\la_p + \ga)^{-1} \dep^2
\leq n \gamma^{-1} \la_{N+1} \sum_{p=1}^{\infty} \la_p^{-1} \dep^2
\leq \gamma^{-1} \la_{N+1}\; n \eta_n^2  \eqsp,
\end{equation}
which goes to zero uniformly in $n$ as $N \to \infty$.
Therefore, under Assumptions (B\ref{hypB:sq_sum_eigs}) and (C),
we may choose $N$ large enough so that
 \begin{equation}
 \label{eq:from_hat_inf_to_hat_trunc}
 | \E[ \rme^{ \rmi u\hV_{n,\infty}(\ga)} ] -\E[ \rme^{\rmi u\hV_{n,N}(\ga)} ] |
< \epsilon \eqsp .
\end{equation}
Similar calculations allow to prove that
$\E[( V_{n,\infty}(\ga) -V_{n,N}(\ga) )^2] = o(1)$,
which yields that for all $\epsilon >0$, for a sufficiently large $N$, we have
 \begin{align}
 \label{eq:from_inf_to_trunc}
 | \E[ \rme^{ \rmi u V_{n,\infty}(\ga)} ] -\E[ \rme^{\rmi u V_{n,N}(\ga)} ] |
< \epsilon \eqsp .
\end{align}
Finally, combining~(\ref{eq:clt_trunc}) and
(\ref{eq:from_hat_inf_to_hat_trunc})~(\ref{eq:from_inf_to_trunc}),
by the triangular inequality, we have proved that, for $\epsilon >0$, we may
choose a sufficiently large $N$, such that
\begin{equation}
 | \E[ \rme^{ \rmi u\hV_{n,\infty}(\ga)} ] - \E[ \rme^{ \rmi u V_{n,\infty}(\ga)} ]  |< \epsilon  \eqsp ,
\end{equation}
and the proof is concluded by invoking L\'{e}vy's continuity theorem~\cite[Theorem 26.3]{Billingsley:1995}.
\end{proof}

\begin{remark}
\label{remark:trunc_neg_relax}
For the sake of generality, we proved the result under the assumption that $n\eta_n^2 =O(1)$.
However, if there exists a nonnegative nondecreasing sequence of integers $\{q_n\}_{n \geq 1}$
such that for all $n$ we have $\sum_{p=1}^{\infty} (\la_p +\ga)^{-1} \pscal{\dan}{e_p}^2 = 
 (\la_{q_n}+\ga)^{-1} \pscal{\dan}{e_{q_n}}^2 $, then the truncation argument used in (\ref{eq:trunc_neg}) is valid 
 under a weaker assumption. In particular, when considering non-directional sequences of 
 local alternatives as in~(\ref{eq:shifting_freq_alt}), it suffices to take $N \to \infty$
 such that $N^{-1}q_n = o(1)$, which for $n$ sufficiently large allows to get 
$
 n \sum_{p=N+1}^{\infty} (\la_p+\ga)^{-1} \pscal{\dan}{e_p}^2 = 0 
$
in place of~(\ref{eq:trunc_neg}) in the proof. The rest of the proof follows similarly.
\end{remark}

The following lemma highlights the main difference between the asymptotics
respectively when \condgaf~and $\ga_n \to 0$, which is that $d_1(\S_1, \ga_n) \to \infty$
and $d_2(\S_1, \ga_n) \to \infty$
in the case $\ga_n \to 0$, whereas they acted
as irrelevant constants in the case \condgaf.
\begin{lem}
\label{lem:c-d-go-to-infinity}
 If $\gamma_n = o(1)$,
 then, $d_{1}(\S_1, \ga_n)\to \infty$, and $d_{2}(\S_1, \ga_n)\to \infty$, as $n$ tends to infinity.
\end{lem}
\begin{proof}
Since the function $x \mapsto x / (x+\gamma_n)$ is monotone increasing, for any $\lambda \geq \gamma_n$,
$\lambda / (\lambda + \gamma_n) \geq 1/2$. Therefore,
\[
\sum_{p=1}^n \frac{\lambda_p(\S_1)}{\lambda_p(\S_1)+\gamma_n} \geq \frac12 \# \left\{ k \leq n : \lambda_p(\S_1) \geq \gamma_n \right\} \eqsp,
\]
and the proof is concluded by noting that since $\gamma_n \to 0$, $\# \left\{ k : \lambda_p(\S_1) \geq \gamma_n \right\} \to \infty$,
as $n$ tends to infinity.
\end{proof}

The quantities $\la_p(\S_1), \la_q(\S_1), d_1(\S_1,\ga_n), d_2(\S_1,\ga_n)$ being pervasive in the subsequent proofs,
they shall be respectively be abbreviated as $\la_p, \la_q, d_{1,n}, d_{2,n}$. Our test statistics writes as
$\tT_n = (\sqrt{2} d_{2,n})^{-1} A_n$ with
\begin{equation}
\label{eq:definition-A}
A_n \eqdef \frac{n_1 n_2}{n} \; \norm{(\S_1 + \gamma_n \Id)^{-1/2} \hat{\delta}}^2 - d_{1,n} \eqsp .\end{equation}
Using the quantities $S_{n,p}$ and $Y_{n,p,i}$
defined respectively in (\ref{eq:definition-Snp}) and (\ref{eq:definition-Ynp}),
$A_n$ may be expressed as
\begin{align*}
A_n &= \sum_{p=1}^\infty \left( \la_p + \gamma_n \right)^{-1}
\left( S_{n,p} + \sqrt{ \frac{n_1 n_2}{n} } \dep \right) ^2 \quad- d_{1,n} \\
&= \sum_{p=1}^\infty  \left( \la_p + \gamma_n \right)^{-1}
\left\{S_{n,p}^2 - \E S_{n,p}^2 + 2 \sqrt{\frac{n_1 n_2}{n}} S_{n,p} \dep \right\}\\
&\quad + \frac{n_1 n_2}{n} \pscal{\dan}{(\S_1 + \ga_n \Id)^{-1} \dan}
+ \sum_{p=1}^\infty \left( \la_p + \gamma_n \right)^{-1} \left\{ \E S_{n,p}^2 - \la_p \right\} \eqsp.
\end{align*}
Since, by Lemma \ref{lem:formulae} Eq.~\eqref{eq:formulae:sum}, $\E S_{n,p}^2- \la_p= (n_1/n) \la_p  \epp$, where $\epp$ is
defined in \eqref{eq:epq_def},  then, by H\"{o}lder inequality, we obtain
\[
\left|\sum_{p=1}^\infty \left( \la_p + \gamma_n \right)^{-1} \left\{ \E S_{n,p}^2 - \la_p\right\}\right|\\
\leq \left( \sum_{p=1}^\infty \left( \la_p + \gamma_n \right)^{-2} \la_p^2 \right)^{1/2}
\left(\sum_{p=1}^\infty  \epp^2 \right)^{1/2} = O(d_{2,n} \eta_n ) \eqsp.
\]
We now decompose
\[
\sum_{p=1}^\infty  \left( \la_p + \gamma_n \right)^{-1}
\left\{ S_{n,p}^2 - \E S_{n,p}^2 + 2 \sqrt{\frac{n_1 n_2}{n}} S_{n,p} \dep \right\} = B_n + 2 C_n + 2 D_n \eqsp,
\]
where $B_n$ and $C_n$ and $D_n$ are defined as follows
\begin{align}
\label{eq:definition-B}
B_n &\eqdef \sum_{p=1}^\infty  \sum_{i=1}^n \left\{ Y_{n,p,i}^2 - \E Y_{n,p,i}^2 \right\} \eqsp, \\
\label{eq:definition-C}
C_n &\eqdef \sum_{p=1}^\infty  \left( \la_p + \gamma_n \right)^{-1} \sum_{i=1}^n Y_{n,p,i} \sqrt{\frac{n_1 n_2}{n}}  \dep   \eqsp,
\\
\label{eq:definition-D}
D_n &\eqdef \sum_{p=1}^\infty  \left( \la_p + \gamma_n \right)^{-1} \sum_{i=1}^n Y_{n,p,i} \left\{ \sum_{j=1}^{i-1}  Y_{n,p,j}  \right\} \eqsp.
\end{align}
The proof is in three steps. We will first show that $B_n$ is negligible, then that $C_n$ is negligible,
and finally establish a central limit theorem for $D_n$.
\begin{proof}[Step 1: $B_n = o_P(1)$]
The proof amounts to compute the variance of this term. Since the variables $Y_{n,p,i}$  and $Y_{n,q,j}$ are independent if $i \neq j$,
then $\PVar(B_n)= \sum_{i=1}^n  v_{n,i}$, where
\begin{align*}
v_{n,i} &\eqdef \PVar \left( \sump (\la_p + \gamma_n)^{-1} \{ Y_{n,p,i}^2 - \E[Y_{n,p,i}^2] \} \right)\\
&= \sumpaq (\la_p + \gamma_n)^{-1} (\la_q + \gamma_n)^{-1} \PCov(Y^2_{n,p,i},Y^2_{n,q,i}) \eqsp.
\end{align*}
Using Lemma~\ref{lem:formulae}, Eq.~\eqref{eq:borne-covariance-carre}, we get
\[
\sum_{i=1}^n v_{n,i}
\leq  C n^{-1}  \left(\sump (\la_p + \gamma_n)^{-1} \la_p^{1/2} (1 + \epp)^{1/2} \right)^2
\leq C n^{-1} \gamma_n^{-2} \left( \sump \la_p^{1/2} \right)^2 \{1+ O(\eta_n)\}
\]
where the RHS above is indeed negligible, since by assumption we have $\ga_n^{-1}n^{-1/2} \to 0$ and  $\sump \la_p^{1/2}< \infty$.
\end{proof}
\begin{proof}[Step 2: $C_n = o_P(d_{2,n}^2)$] Again, the proof essentially consists in computing
the variance of this term, and then conclude by the Markov inequality. As previously,
since the variables $Y_{n,p,i}$  and $Y_{n,q,j}$ are independent if $i \neq j$,
then $\PVar(C_n)= \sum_{i=1}^n  u_{n,i}$, where
\begin{multline*}
u_{n,i} \eqdef \sump (\la_p + \gamma_n)^{-2}  \E [Y_{n,p,i}^2] \frac{n_1 n_2}{n} \dep^2 \\
+ \sumpaq  (\la_q + \gamma_n)^{-1} (\la_q + \gamma_n)^{-1} \E [Y_{n,p,i} Y_{n,q,i}] \frac{n_1 n_2}{n} \dep \deq
\eqsp.
\end{multline*}
Moreover, note that $\E [Y_{n,p,i}^2] \leq C n^{-1} \la_p$, and under Assumption~(C\ref{hypC})
\begin{multline*}
\pons d_{2,n}^{-2} \sump (\la_p + \ga_n)^{-2} \la_p \dep^2 \\
= \pons d_{2,n}^{-2} \sump (\la_p + \ga_n)^{-1}  \dep^2
\leq d_{2,n}^{-1} \frac{n \pscal{\dan}{(\S_1 + \ga_n)^{-1}\dan} }{d_{2,n}} =o(1) \eqsp .
\end{multline*}
Similarly, for $p \neq q$ we have $|\E [Y_{n,p,i} Y_{n,q,i}]|\leq C n^{-1} \la_p^{1/2} \la_q^{1/2} |\epq|$,
which implies that
\begin{multline*}
\pons d_{2,n}^{-2} \sumpq (\la_p + \ga_n)^{-1} (\la_q + \ga_n)^{-1} \la_p^{1/2} \la_q^{1/2} |\dep| |\deq| |\epq| \\
\leq \pons d_{2,n}^{-2} \left(\sump  (\la_p + \ga_n)^{-2} \la_p \dep^2 \right) \left(\sumpq \epq^2 \right)^{1/2} = o(1)  \eqsp .
\end{multline*}
\end{proof}

\begin{proof}[Step 3:  $d_{2,n}^{-1} D_{n}  \dlim \gauss(0,1/2)$]
We use the central limit theorem (CLT) for triangular array of martingale difference~\citep[Theorem 3.2]{Hall:Heyde:1980}.
For $=1,\dots, n$, denote
\begin{equation}
\label{eq:definition-xi-2}
\xi_{n,i} \eqdef d^{-1}_{2,n} \sump (\la_p+\gamma_n)^{-1}   Y_{n,p,i} M_{n,p,i-1} \eqsp,
\quad \text{where} \quad M_{n,p,i} \eqdef \sum_{j=1}^{i} Y_{n,p,j} \eqsp,
\end{equation}
and let  $\mcf_{n,i}= \sigma\left(  Y_{n,p,j}, p \in \{1, \dots, n\}, j \in \{0, \dots, i\} \right)$.
Note that, by construction, $\xi_{n,i}$ is a martingale increment, that is~$\CPE{\xi_{n,i}}{\mcf_{n,i-1}}= 0$.
The first step in the proof of the CLT is to establish that
\begin{equation}
\label{eq:definition-sn2}
s_n^2=\sum_{i=1}^n \CPE{\xi_{n,i}^2}{\mcf_{n,i-1}} \plim 1/2 \eqsp.
\end{equation}
The second step of the proof is to establish the negligibility condition. We invoke~\citep[Theorem 3.2]{Hall:Heyde:1980}, which requires to establish that
$\max_{1 \leq i \leq n} |\xi_{n,i} | \plim 0$ (smallness) and $\E (\max_{1 \leq i \leq n} \xi_{n,i}^2)$ is bounded in $n$ (tightness), where
$\xi_{n,i}$ is defined in \eqref{eq:definition-xi-2}. We will establish the two conditions simultaneously by checking that
\begin{equation}
\label{eq:negligibility}
\E \left(\max_{1 \leq i \leq n} \xi_{n,i}^2 \right)= o(1) \eqsp.
\end{equation}
Splitting the sum $s_n^2$,
between diagonal terms $E_n$,
and off-diagonal terms $F_n$,
we have
\begin{align}
\label{eq:definition-E_n}
E_n &= d_{2,n}^{-2}\sump (\la_p+\ga_n)^{-2} \sumi M_{n,p,i-1}^2 \E [Y_{n,p,i}^2] \eqsp, \\
\label{eq:definition-F_n}
F_n &= d_{2,n}^{-2}\sumpq (\la_p+\ga_n)^{-1} (\la_q+\ga_n)^{-1}
\sumi M_{n,p,i-1} N_{n,q,i-1} \E [Y_{n,p,i} Y_{n,q,i}]\eqsp.
\end{align}
Consider first the diagonal terms $E_n$.  We first compute its mean. Note that $\E[N^2_{n,p,i}] = \sum_{j=1}^i \E[Y_{n,p,j}^2]$.
Using Lemma~\ref{lem:formulae}, we get
\begin{multline*}
\sump (\la_p+\ga_n)^{-2} \sumi \sum_{j=1}^{i-1} \E [ Y^2_{n,p,j}] \E [ Y_{n,p,i}^2] \\
= \frac{1}{2} \sump (\la_p+\ga_n)^{-2} \left\{ \left[ \sum_{i=1}^n \E[ Y_{n,p,i}^2] \right]^2 - \sum_{i=1}^n \E^2[Y^2_{n,p,i}] \right\}
= \frac{1}{2} d_{2,n}^2 \left\{1 + O(d_{2,n}^{-1}\eta_n)  + O(n^{-1})\right\} \eqsp.
\end{multline*}
Therefore, $\E[E_n]= 1/2+o(1)$.  Next, we check that $E_n-\E[E_n]= o_P(1)$ is negligible.
We write $E_n - \E[E_n]= d_{2,n}^{-2} \sum_{p=1}^n (\la_p + \gamma_n)^{-2} Q_{n,p}$, with
\begin{equation}
\label{eq:definition-Qnp}
Q_{n,p} \eqdef \sum_{i=1}^n
 \E[ Y_{n,p,i+1}^2 ] \left\{ N^2_{n,p,i} - \E \left[ N^2_{n,p,i} \right] \right\} \eqsp.
\end{equation}
Using this notation,
\begin{multline}
\label{eq:expression-variance-Dn}
\PVar[E_n] = d_{2,n}^{-4}\sum_{p=1}^n (\la_p + \gamma_n)^{-4} \E[ Q^2_{n,p}] \\
+ 2 d_{2,n}^{-4} \sum_{1 \leq p < q \leq n} (\la_p + \gamma_n)^{-2} (\la_q + \gamma_n)^{-2} \E[ Q_{n,p} Q_{n,q} ] \eqsp.
\end{multline}
We will establish that
\begin{equation}
%\label{eq:borne_Qnp^2}
%&\E [Q_{n,p}^2 ]
%\leq C \left\{\lambda_p^4 + n^{-1}\lambda_p^3 \right\}\\
\label{eq:borne_QnpQnq}
\left|\E [Q_{n,p} Q_{n,q}]\right|
\leq C \left\{\lambda_p^2 \lambda_q^2 (\delta_{p,q} + |\epq|)^2 + n^{-1}\lambda_p^{3/2} \lambda_q^{3/2}  \right\} \eqsp .
\end{equation}
Plugging this bound into \eqref{eq:expression-variance-Dn} and using that $\lambda_p/ (\lambda_p + \gamma_n) \leq 1$ and $d_{2,n} \to \infty$ as $n$ tends to infinity, yields under Assumption~(B\ref{hypB:sq_sum_eigs})
\begin{align*}
\PVar[E_n]
&\leq
\left\{
 d_{2,n}^{-2}
+ n^{-1} \ga_n^{-1} d_{2,n}^{-2}
\right\}
+ C
\left\{
 d_{2,n}^{-2}    \eta_n
+ n^{-1} d_{2,n}^{-4} \; \left(\sump \la_p \right)^{2}
\right\} \eqsp,
\end{align*}
showing that $\PVar[E_n]= o(1)$, and hence that $E_n - \E[E_n]= o_P(1)$. To show  \eqref{eq:borne_QnpQnq},  note first
that $\{ M_{n,p,i}^2- \E [ M_{n,p,i}^2] \}_{1 \leq i \leq n}$ is a $\mcf_{n}$-adapted martingale.
Denote by $\nu_{n,p,i}$ its increment defined recursively as follows:
$\nu_{n,p,1}= N_{n,p,1}^2- \E [ N_{n,p,1}^2]$ and for $i \geq 1$ as
\[
\nu_{n,p,i} = M_{n,p,i}^2- \E [ M_{n,p,i}^2] - \left\{ N^2_{n,p,i-1} - \E[ N^2_{n,p,i-1} ] \right\} =
Y_{n,p,i}^2 - \E [Y_{n,p,i}^2] + 2 Y_{n,p,i} M_{n,p,i-1} \eqsp.
\]
Using the summation by part formula, $Q_{n,p}$ may be expressed as
\[
Q_{n,p}= \sum_{i=1}^{n-1} \nu_{n,p,i} \left[ \sum_{j=i+1}^n \E[ Y_{n,p,j}^2] \right] \eqsp.
\]
Using Lemma~\ref{lem:formulae}, Eq.~\eqref{eq:formulae:sum}, we  obtain for any $1 \leq p \leq q \leq n$,
\begin{align}
\left|\E[ Q_{n,p} Q_{n,q}] \right| &\leq \left( \sum_{j=1}^n \E[ Y_{n,p,j}^2] \right) \left( \sum_{j=1}^n \E[ Y_{n,q,j}^2] \right) \left|\sum_{i=1}^{n-1}\E[ \nu_{n,p,i} \nu_{n,q,i} ] \right| \nonumber \\
&\leq C \lambda_p  \lambda_q  ( 1 + O(\eta_n)) \label{eq:covQnpq}
\left|\sum_{i=1}^{n-1} \E[ \nu_{n,p,i} \nu_{n,q,i} ] \right|\eqsp.
\end{align}
We get
\[
\E[ \nu_{n,p,i} \nu_{n,q,i} ]= \PCov(Y_{n,p,i}^2,Y_{n,q,i}^2) + 4 \E\left\{ Y_{n,p,i}Y_{n,q,i} \right\} \E\left\{M_{n,p,i-1} N_{n,q,i-1}\right\}\eqsp.
\]
First, applying Eq.~\eqref{eq:borne-covariance-carre} in Lemma~\ref{lem:formulae} gives
\begin{equation}
\label{eq:borne-carre-covariance}
\sum_{i=1}^{n-1} \PCov(Y_{n,p,i}^2,Y_{n,q,i}^2)
\leq C n^{-1} \la_p^{1/2}  \la_q^{1/2}  \eqsp .
\end{equation}
Since $\E[M_{n,p,i-1} N_{n,q,i-1}]= \sum_{j=1}^{i-1} \E [ Y_{n,p,j}Y_{n,q,j}]$, Lemma~\ref{lem:formulae}, Eq.~\eqref{eq:formulae:sum}  shows that
\begin{align}
\left|\sumi \E [ Y_{n,p,i}Y_{n,q,i}]\E[M_{n,p,i-1} N_{n,q,i-1}]\right|
&= \frac{1}{2} \left|\left\{ \left( \sum_{i=1}^n \E[ Y_{n,p,i}Y_{n,q,i}] \right)^2 - \sum_{i=1}^n \E^2[Y_{n,p,i}Y_{n,q,i}] \right\}\right| \nonumber \\ \label{eq:prev}
&\leq C \lambda_p  \lambda_q  (\delta_{p,q} + |\epq|)^2 \eqsp .
\end{align}
Eq.~\ref{eq:borne_QnpQnq} follows by plugging \eqref{eq:borne-carre-covariance} and \eqref{eq:prev} into \eqref{eq:covQnpq}.
We finally consider $F_n$ defined in \eqref{eq:definition-F_n}.
We will establish that $F_n = o_P(1)$. Using Lemma~\ref{lem:formulae}-Eq.~\eqref{eq:formulae:sum},
\[
\E^{1/2}[M_{n,p,i-1}^2] \E^{1/2}[N_{n,q,i-1}^2] \leq C \la^{1/2}_p \la^{1/2}_q \eqsp,
\]
and $|\E[ Y_{n,p,i} Y_{n,q,i}]| \leq C n^{-1} \la_p^{1/2} \la_q^{1/2} \epq$, the Minskovski inequality implies that
\[
\{ \E |F_n|^2 \}^{1/2} \leq C d_{2,n}^{-2}\sumpq (\la_p+\ga_n)^{-1} (\la_q+\ga_n)^{-1} \la_p \la_q \epq \leq
C \eta_n \eqsp,
\]
showing that $F_n = o(1)$. This concludes the proof of Eq.~\eqref{eq:definition-sn2}.

We finally show Eq.~\eqref{eq:negligibility}. Since $|Y_{n,p,i}| \leq n^{-1/2} |k|_\infty^{1/2}$ $\P$-\as\, we may bound
\begin{equation}
\label{eq:maxou}
\max_{1 \leq i \leq n} |\xi_{n,i}|
\leq C d_{2,n}^{-1} n^{-1/2} \sump (\la_p + \ga_n)^{-1} \max_{1 \leq i \leq n} |M_{n,p,i-1}| \eqsp .
\end{equation}
Then, the Doob inequality implies that
 $
\E^{1/2}[ \max_{1 \leq i \leq n} |M_{n,p,i-1}|^2] \leq  \E^{1/2}[ N^2_{n,p,n-1} ] \leq C \la_p^{1/2}\eqsp.
$
Plugging this bound in  \eqref{eq:maxou}, the Minkowski inequality
\begin{equation*}
\E^{1/2} \left( \max_{1 \leq i \leq n} \xi_{n,i}^2 \right)
\leq C \left\{d_{2,n}^{-1} \ga_n^{-1} n^{-1/2} \sump \la_p^{1/2} \right\} \eqsp ,
\end{equation*}
and the proof is concluded using the fact that \condgan\ and Assumption~(B\ref{hypB:sq_sum_eigs}).
\end{proof}

%----------------------------------------------------------------------------------------------------------------------------------------

\section{Proof of Theorem \ref{theo:test_power_consistency}}

\begin{proof}[Proof of Proposition \ref{prop:imagekernel}]
We denote by $\S = \rho_1 \S_1 + \rho_2 \S_2 + \rho_1 \rho_2 \delta \otimes \delta$  the covariance operator associated with the probability density $p = \rho_1 p_1 + \rho_2 p_2$, 
and $\delta = \mu_2 - \mu_1$. Then, Proposition~\ref{prop:range} applied to the probability densities $p_1,p_2$ and $p= \rho_1 p_1 + \rho_2 p_2$ shows
that $\pscal{\delta}{\S^{-1} \delta}_{\H}  = \int \frac{ (p_1 - p_2)^2}{\rho_1 p_1 + \rho_2 p_2} d \rho$.
Thus
\begin{align*}
\rho_1 \rho_2 \pscal{\delta}{\S^{-1} \delta}_{\H}  %\leq \rho_1 \rho_2 \int \frac{ (p_1 - p_2)^2}{\rho_1 p_1 + \rho_2 p_2}    \\
&= \frac{1}{2} \int \frac{ \frac{\rho_1}{\rho_2} (p_1 - p)^2 + \frac{\rho_2}{\rho_1} (p_2 - p)^2 }{p} d \rho\\
%& = & \frac{1}{2} \frac{\rho_1}{\rho_2} \left( \int \frac{p_1^2}{p} - 1 \right) + \frac{1}{2} \frac{\rho_2}{\rho_1} \left( \int \frac{p_2^2}{p} - 1 \right)
&=  \frac{1}{2 \rho_1 \rho_2 } \int \frac{ \rho_1^2 p_1^2 + \rho_2^2 p_2^2 }{ p} d \rho - \frac{1}{2} \frac{\rho_2}{\rho_1} -  \frac{1}{2} \frac{\rho_1}{\rho_2} \\
&=  \frac{1}{2 \rho_1 \rho_2 }  - \frac{1}{2} \frac{\rho_2}{\rho_1} - \frac{1}{2} \frac{\rho_1}{\rho_2} -  \int \frac{ p_1 p_2 } {p } d \rho
%= \frac{ 1- \rho_1^2 - \rho_2^2 }{2 \rho_1 \rho_2 } -  \int \frac{ p_1 p_2 } {p }
= 1 -   \int \frac{ p_1 p_2 } {p } d \rho \eqsp.
\end{align*}
The previous inequality shows that $\rho_1 \rho_2\pscal{\delta}{\S^{-1} \delta}_{\H}<1$ is satisfied when $\int p_1 p_2 / p d \rho \ne 0$.
Therefore, in this situation,
\begin{multline*}
\pscal{\delta}{( \rho_1 \S_1 + \rho_2 \S_2 )^{-1} \delta}_{\H} =
\pscal{\delta}{( \S - \rho_1 \rho_2 \delta \otimes \delta   )^{-1}  \delta}_{\H} \\=
\pscal{\delta}{ \S^{-1} \delta}_{\H} (  1 - \rho_1 \rho_2 \pscal{\delta}{\S^{-1} \delta}_{\H})^{-1} \eqsp,
\end{multline*}
and the proof follows by combining the two latter equations.

Consider now the case where $\int p_1 p_2/ p  d \rho= 0$, that is when the probability distribution $\P_1$ and $\P_2$ are
singular (for any set $A \in \Xsigma$ such as $\P_1(A) \ne 0$, $\P_2(A) = 0$ and vice-versa).
In that case, $\pscal{\delta}{( \rho_1 \S_1 + \rho_2 \S_2 )^{-1} \delta}_{\H}$ is infinite.
\end{proof}

\begin{proof}
We first prove that
\begin{equation}
\label{eq:conv_prob_maha}
\norm{(\hS_{W} + \gamma_n \Id)^{-1/2} \hat{\delta}}_{\H}^2 \plim \norm{(\S_{W} + \gamma_\infty I)^{-1/2} {\delta}}_{\H}^2 \eqsp,
\end{equation}
where $\gamma_\infty \eqdef \lim_{n \to \infty} \gamma_n$. Using straightforward algebra, we may write
\begin{equation}
\label{eq:C_decomp}
\left|\norm{(\hS_{W} + \gamma_n \Id)^{-1/2} \hat{\delta}}_{\H}^2
-\norm{(\S_{W} + \gamma_\infty I)^{-1/2} {\delta}}_{\H}^2\right|
 \leq C_1 + C_2 + C_3 \eqsp ,
\end{equation}
where
\begin{align*}
\nonumber
C_1 &\eqdef \norm{({\S}_{W} + \gamma_n \Id)^{-1/2} \hat{\delta}}_{\H}
\norm{({\hS}_{W} + \gamma_n \Id)^{-1/2} \hat{\delta}}_{\H} \hsnorm{( \hS_W + \gamma_n \Id)^{-1/2} (\hS_W - \S_W) \left( \S_W + \gamma_n \Id  \right)^{-1/2}}\eqsp , \\
\nonumber
C_2 &\eqdef \left|\left( \norm{({\S}_{W}+ \gamma_n \Id)^{-1/2} \hat{\delta}}_{\H}^2 - \norm{ (\S_{W} + \gamma_n \Id)^{-1/2} {\delta}}^2_{\H}  \right) \right| \eqsp ,\\
\nonumber
C_3 &\eqdef \left|\left( \norm{(\S_{W} + \gamma_n \Id)^{-1/2} {\delta}}_{\H}^2 - \norm{( \S_{W} + \gamma_\infty I)^{-1/2} {\delta}}_{\H}^2  \right)\right| \eqsp .
\end{align*}
First, prove that $C_1 = o_P(1)$. Write $C_1 = A_1 A_2 B_1$.
Using (with obvious changes) the relation \eqref{eq:zaid}, the monotone convergence theorem yields
\begin{equation*}
\label{eq:boundedness-in-norm}
\lim_{n \to \infty} \E \norm{({\S}_{W} + \gamma_n \Id)^{-1/2} \hat{\delta}}_{\H}^2
=\pscal{ \delta}{(\S_W + \gamma_\infty I)^{-1} \delta}_{\H} \eqsp.
\end{equation*}
which gives  $A_1 = O_P(1)$.
As for proving $A_2 =  O_P(1)$, using an argument similar to the one used to derive Eq.~\eqref{eq:A2_eq},
it suffices to observe that $A_2 = A_1 + o_P(1)$. Then,
Eq.~\eqref{eq:hs_conv_cov_op_eq} in Corollary~\ref{cor:hs_conv_cov_op} gives $B_1 = O_P(\ga_n^{-1}n^{-1/2})$,
which shows that $C_1 =A_1 A_2 B_1 = o_P(1)$.
Next, prove that $C_2 = o_P(1)$.
We may write
\begin{equation*}
C_2 = 2 \pscal{\hat{\delta}-\delta}{(\S_W + \gamma_n \Id)^{-1} \delta}_{\H} + \norm{(\S_W + \gamma_n \Id)^{-1/2} (\hat{\delta}-\delta)}^2_{\H} \eqsp.
\end{equation*}
Since $\openorm{(\S_W + \ga_n \Id)^{-1/2}}_{\H} \leq \ga_n^{-1/2}$, and $\norm{(\S_W + \ga_n \Id)^{-1/2} \delta}_{\H} < \infty$,
and moreover $\Vert \hat{\delta}-\delta \Vert_{\H} = O_P(n^{-1/2})$, then we get $C_2= O_P(\ga_n^{-1/2} n^{-1/2})=o_P(1)$
Finally, prove that $C_3 = o(1)$
Note that $C_3 = - \sum_{p=1}^\infty \ga_n^{-1} (\la_p + \ga_n)^{-1} \la_p \pscal{\delta}{e_p}^2_{\H}$,
where $\{\la_p\}$ and $\{e_p\}$ denote respectively the eigenvalues and eigenvectors of $\S_W$.
Since $[\ga \mapsto (\la_p + \ga)^{-1} \ga]$ is monotone, the monotone convergence theorem shows that $C_3= o(1)$.

Now, when $\P_1 \neq \P_2$, Proposition~\ref{prop:imagekernel} with $\P = \rho_1 \P_1 + \rho_2 \P_2$
ensures that $\delta \in \mathcal{R}(\S_W^{1/2})$ as long as $\norm{\frac{d \P_2}{d \P_1} - 1}_{\ltwo(\P_1)} < \infty$.
Then, under assumption (A\ref{hypA:Kernel:injection}), by injectivity of $\S_W$ we have $\delta \neq 0$.
Hence, since $\S_W$ is trace-class, we may apply Lemma~\ref{lem:bornesuite} with $\alpha =1$,
which yields $d^{-1}(\S_W,\gamma_n) \; n \to \infty$. Therefore, $\hT_n(\ga_n) \plim \infty$,
and the proof ois concluded. Otherwise, that is
when $\norm{\frac{d \P_2}{d \P_1} - 1}_{\ltwo(\P_1)} = \infty$, we have $\hT_n(\ga_n) \plim \infty$.
\end{proof}
%%%%%%%%%%%%%%%%%%%%%%%%%%%%%%%%%%%%%%%%%%%%%%%%%%%%%%%%%%%%%%%%%%%%%%%%%%%%%%%%%%%%%%%%%%%%%%%%%%%%%%%%%%%%%%%%%%%%%%%%%%%%%%%%%%%%%%%%%

%%%%%%%%%%%%%%%%%%%%%%%%%%%%%%%%%%%%%%%%%%%%%%%%%%%%%%%%%%%%%%%%%%%%%%%%%%%%%%%%%%%%%%%%%%%%%%%%%%%%%%%%%%%%%%%%%%%%%%%%%%%%%%%%%%%%%%%%%
\appendix
\section{Technical Lemmas}
\begin{lem}
\label{lem:bornesuite}
Let $\{\lambda_p \}_{p \geq 1}$ be a non-increasing sequence of non-negative numbers. Let $\alpha > 0$. Assume that $\sum_{p \geq 1} \lambda_p^\alpha < \infty$. Then, for
any $\beta \geq \alpha$,
\begin{equation}
\label{eq:bornesuite-1}
\sup_{\gamma > 0} \gamma^\alpha \sum_{p=1}^\infty \lambda_p^\beta (\lambda_p + \gamma)^{-\beta} \leq 2 \sum_{p=1}^\infty \lambda_p^\alpha \eqsp.
\end{equation}
In addition, if $\lim_{p \to \infty} p \lambda_p^\alpha = \infty$, then for any $\beta > 0$,
\begin{equation}
\label{eq:bornesuite-3}
\lim_{\gamma \to 0} \gamma^\alpha \sum_{p=1}^\infty \lambda_p^\beta (\lambda_p + \gamma)^{-\beta}= \infty \eqsp.
\end{equation}
\end{lem}
\begin{proof}
For $\gamma > 0$, denote by $q_\gamma = \sup_{p \geq 1} \{ p: \lambda_p > \gamma \}$. Then,
\begin{equation}
\label{eq:zebound}
\gamma^\alpha \sum_{p=1}^\infty \lambda_p^\beta (\lambda_p + \gamma)^{-\beta} \leq  \gamma^{\alpha} \sum_{p=1}^\infty \lambda_p^\alpha (\lambda_p + \gamma)^{-\alpha} \leq \gamma^{\alpha} q_\gamma +  \sum_{p > q_\gamma}^\infty \lambda_p^\alpha \eqsp.
\end{equation}
Since the sequence $\{ \lambda_p \}$ is non-increasing, the condition $C \eqdef \sum_{p \geq 1} \lambda_p^\alpha < \infty < \infty$ implies that
$p \lambda_p^\alpha \leq C$. Therefore, $\lambda_p \leq C^{1/\alpha} p ^{-1/\alpha}$, which implies that for any $p$ satisfying $C \gamma^{-\alpha} \leq p$,
$\lambda_p \leq \gamma$, showing that $q_\gamma \leq C \gamma^{-\alpha}$. This establishes \eqref{eq:bornesuite-1}.

Since $\lambda \mapsto \lambda (\lambda + \gamma)^{-1}$ is non-decreasing, for $p \leq q_\gamma$, $\lambda_p (\lambda_p+\gamma)^{-1} \geq  (1/2)$. Therefore,
$\gamma^\alpha \sum_{p=1}^\infty \lambda^\beta_p (\lambda_p + \gamma)^{-\beta} \geq (2)^{-\beta} \gamma^\alpha q_\gamma$. Since $\lim_{ p \to \infty} p \lambda_p^\alpha= \infty$,
this means that $\lambda_p > 0$ for any $p$, which implies that $\lim_{\gamma \to 0^+} q_\gamma= \infty$. Therefore, $\lim_{\gamma \to 0^+} q_\gamma \lambda_{q_\gamma}^\alpha = \lim_{\gamma \to 0^+} q_\gamma \gamma^\alpha = \infty$.
The proof follows.
\end{proof}

\begin{lem}
\label{lem:d_equiv}
Let $\{\lambda_p \}_{p \geq 1}$ be a non-increasing sequence of non-negative numbers.
Assuse there exists $s >0$ such that $\la_p = p^{-s}$ for all $p \geq 1$.  Then,
\begin{equation*}
\left[
\sum_{p=1}^{\infty} (\la_p + \ga_n)^{-r} \la_p^r
\right]^{1/r}
= \ga^{-1/{s}{r}} \left\{ \int_{0}^{\infty} (1+v^{s})^{-r} dv \right\}^{1/r} (1+o(1))
\eqsp,\quad\text{as}\;\ga \to 0 \eqsp .
\end{equation*}
\end{lem}

\begin{proof}
First note that
\begin{equation*}
\sump (\la_p + \ga_n)^{-r} \la_p^r = \sump (1 + \ga_n\la_p^{-1})^{-r} 
= \sump (1 + (\ga_n^{1/s} p)^{s})^{-r}  \eqsp .\\
\end{equation*}
For all $\ga >0$, the function $[u \mapsto (1+(\ga^{1/s} u)^{s})^{-r}]$ is increasing and nonnegative. Therefore, for all $p \geq 1$ we may write
\begin{eqnarray*}
\int_{p}^{p+1} (1+(\ga^{1/s} u)^{s})^{-r} du &\leq (1+(\ga^{1/s} p)^{s})^{-r} &\leq \int_{p-1}^{p} (1+(\ga^{1/s} u)^{s})^{-r} du \eqsp , \\
\ga^{-1/s} \int_{\ga^{1/s} p}^{\ga^{1/s}(p+1)} (1+v^{s})^{-r} dv &\leq (1+(\ga^{1/s} p)^{s})^{-r} &\leq \ga^{-1/s} \int_{\ga^{1/s}(p-1)}^{\ga^{1/s} p} (1+ v^{s})^{-r} dv  \eqsp .
\end{eqnarray*}
Hence, sussing on $p$ over $1,\dots, N-1$, we obtain
\begin{eqnarray*}
\ga^{-1/s} \int_{\ga^{1/s} }^{\ga^{1/s} N} (1+v^{s})^{-r} dv &\leq \sum_{p=1}^{N} (1+(\ga^{1/s} p)^{s})^{-r} &\leq \ga^{-1/s} \int_{0}^{\ga^{1/s} N } (1+ v^{s})^{-r} dv  \eqsp . \\
\end{eqnarray*}
Therefore, taking $N \to \infty$ in such a way that $\ga^{1/s} N \to \infty$ as $\ga \to 0$, we finally get
\begin{equation*}
\sum_{p=1}^{\infty} (1+(\ga^{1/s} p)^{s})^{-r} = \ga^{-1/s} \left\{ \int_{0}^{\infty} (1+v^{s})^{-r} dv \right\} (1+o(1)) \eqsp . \qedhere
\end{equation*}
\end{proof}

\begin{lem}
\label{lem:borne-sum-vp}
Let $A$ be a self-adjoint compact operator on $\H$. Then, for any orthonormal basis $\{\varphi_p\}_{p \geq 1}$ of $\H$,
\[
\sum_{p=1}^\infty |\lambda_p(A)| \leq \sum_{p=1}^\infty \norm{A \varphi_p}_{\H} \eqsp.
\]
\end{lem}
\begin{proof}
Let $\{\psi_p\}_{p \geq 1}$ be an orthonormal basis of $\H$ consisting of a sequence of eigenvectors of $A$ corresponding
to the eigenvalues $\{ \lambda_p(A) \}$ of this latter operator, so that $\pscal{\psi_p}{A \psi_p}_{\H}= \lambda_p(A)$. Then,
\begin{multline*}
\sum_{p=1}^\infty |\lambda_p(A)| = \sum_{p=1}^\infty \left| \pscal{\psi_p}{A \psi_p}_{\H} \right|
\leq \sum_{q=1}^\infty \sum_{p=1}^\infty \left| \pscal{A \varphi_q}{\psi_p}_{\H} \right| \left| \pscal{\varphi_q}{\psi_p}_{\H}\right| \\
\leq \sum_{q=1}^\infty \left( \sum_{p=1}^\infty \left| \pscal{A \varphi_q}{\psi_p}_{\H} \right|^2 \right)^{1/2} \left( \sum_{p=1}^\infty \left| \pscal{\varphi_q}{\psi_p}_{\H}\right|^2 \right)^{1/2} \leq \sum_{q=1}^\infty \norm{A \varphi_q}_{\H} \eqsp.
\end{multline*}
\end{proof}

%%%%%%%%%%%%%%%%%%%%%%%%%%%%%%%%%%%%%%%%%%%%%%%%%%%%%%%%%%%%%%%%%%%%%%%%%%%%%%%%%%%%%%%%%%%%%%%%%%%%%%%%%%%%%%%%%%%%%%%%%%%%%%%%%%%%%%%%%

%%%%%%%%%%%%%%%%%%%%%%%%%%%%%%%%%%%%%%%%%%%%%%%%%%%%%%%%%%%%%%%%%%%%%%%%%%%%%%%%%%%%%%%%%%%%%%%%%%%%%%%%%%%%%%%%%%%%%%%%%%%%%%%%%%%%%%%%%%
\section{Perturbation results on covariance operators}

\begin{lem}
\label{lem:eric-s-trick}
Let A be a compact self-adjoint operator, with
 $\{\la_p \}_{p \geq 1}$  the eigenvalues of A,
 and $\{e_p \}_{p \geq 1}$ an orthonormal system of eigenvectors of A.
Then, for all integer $k >1$, using the convention $p_{k+1}=p_1$,
\begin{equation*}
\sum_{p=1}^{\infty} \pscal{e_p}{(AB)^k e_p}
= \sum_{p_1=1}^{\infty} \sum_{p_2=1}^{\infty} \dots \sum_{p_k=1}^{\infty}
\left\{ \left(\prod_{j=1}^k \la_{p_j}\right) \left(\prod_{j=1}^k \pscal{e_{p_j}}{B  e_{p_{j+1}}}\right) \right\}
\eqsp .
\end{equation*}
\end{lem}

\begin{proof}
Let $k$ be some integer, fixed throughout the proof. The proof is by induction,
that is, we shall prove that, for all $\ell \in \{1, \dots, k\}$,
\begin{multline*}
\sum_{p=1}^{\infty} \pscal{e_p}{(AB)^k e_p} \\
= \sum_{p_1=1}^{\infty} \sum_{p_2=1}^{\infty} \dots \sum_{p_{\ell}=1}^{\infty}
\left\{ \left(\prod_{j=1}^{\ell -1} \la_{p_j}\right) \left(\prod_{j=1}^{\ell-1}
 \pscal{e_{p_j}}{B  e_{p_{j+1}}}\right)  \pscal{e_{p_{\ell}}}{(AB)^{k- \ell +1}  e_{p_{1}}} \right\},
 \quad \mathcal{P}(\ell) \eqsp .
\end{multline*}

First, for $\ell = 2$, using that $A^{*}e_{p_1} =A e_{p_1} = \la_{p_1} e_{p_1}$,
and $B^{*}e_{p_1} = \sum_{p_2}^{\infty} \pscal{e_{p_1}}{B e_{p_2}}e_{p_2}$,
we indeed have
\begin{align*}
\sum_{p_1=1}^{\infty} \pscal{e_{p_1}}{AB (AB)^{k-1} e_{p_1}}
&= \sum_{p_1=1}^{\infty}  \la_{p_1} \pscal{B^{*}e_{p_1}}{(AB)^{k-1} e_{p_1}} \\
&= \sum_{p_1=1}^{\infty}  \la_{p_1}
\pscal{\sum_{p_2}^{\infty} \pscal{e_{p_1}}{B e_{p_2}}e_{p_2}}{(AB)^{k-1} e_{p_1}} \\
&= \sum_{p_1=1}^{\infty}
 \sum_{p_2=1}^{\infty}   \la_{p_1} \pscal{e_{p_1}}{B e_{p_2}}  \pscal{e_{p_2}}{(AB)^{k-1} e_{p_1}} ,
 \quad \mathcal{P}(2) \eqsp .
\end{align*}

Assume the statement $\mathcal{P}(\ell)$ is true, with $\ell < k-1$.
Let us now marginalize out, first $A$ then $B$ in $(AB)^{k- \ell +1}$,
for the $(\ell +1)$-th time, by summing over an index $p_{\ell +1}$.
Using the same arguments as above, that is $A^{*}e_{p_\ell} = \la_{p_\ell} e_{p_\ell}$
and
$B^{*}e_{p_\ell} = \sum_{p_{\ell +1}}^{\infty} \pscal{e_{p_\ell}}{B e_{p_{\ell+1}}}e_{p_{\ell+1}}$,
\begin{eqnarray*}
& &\sum_{p=1}^{\infty} \pscal{e_p}{(AB)^k e_p}\\
&=& \sum_{p_1=1}^{\infty}\dots \sum_{p_{\ell}=1}^{\infty}
\left\{ \left(\prod_{j=1}^{\ell -1} \la_{p_j}\right) \left(\prod_{j=1}^{\ell -1}
 \pscal{e_{p_j}}{B  e_{p_{j+1}}}\right)  \pscal{e_{p_{\ell}}}{AB (AB)^{k- \ell}  e_{p_{1}}} \right\}\\
&=& \sum_{p_1=1}^{\infty} \dots \sum_{p_{\ell}=1}^{\infty}
\left\{ \left(\prod_{j=1}^{\ell -1} \la_{p_j}\right) \la_{p_{\ell}} \left(\prod_{j=1}^{\ell-1}
 \pscal{e_{p_j}}{B  e_{p_{j+1}}}\right)
    \pscal{B^{*}e_{p_{\ell}}}{(AB)^{k- \ell}  e_{p_{1}}} \right\}\\
&=&     \sum_{p_1=1}^{\infty} \dots \sum_{p_{\ell}=1}^{\infty}  \sum_{p_{\ell +1}=1}^{\infty}
\left\{ \left(\prod_{j=1}^{\ell} \la_{p_j}\right)  \left(\prod_{j=1}^{\ell -1}
 \pscal{e_{p_j}}{B  e_{p_{j+1}}}\right)
  \pscal{e_{p_{\ell}}}{B e_{p_{\ell+1}}}
    \pscal{e_{p_{\ell +1}}}{(AB)^{k- \ell}  e_{p_{1}}} \right\}
     \eqsp ,
     \\
    \end{eqnarray*}
    which proves $\mathcal{P}(\ell +1)$.

    The proof is concluded by a $k$-step induction,
    that is once $A$ in $(AB)^{k}$ is eventually marginalized out $k$-times
    and only the last term $\pscal{e_{p_{k}}}{B  e_{p_{1}}}$ remains.
\end{proof}

\begin{lem}
\label{lem:perturb_cov}
Let $\ga >0$, and $S$ a trace-class operator.
Denote $\{\la_p\}_{p \geq 1}$ and $\{e_p\}_{p \geq 1}$ respectively
the positive eigenvalues and the corresponding eigenvectors of $S$.
Consider $d_{r}(T, \ga)$ for $r=1,2$, with $T$ a compact operator,
as defined in (\ref{eq:definition-c-d}).
If $\Delta$ is a trace-class perturbation operator such that
$\norm{(S + \ga \Id)^{-1}\Delta } < 1$,
and $\carlnorm{\Delta} =\sum_{p=1}^{\infty} \norm{\Delta e_p} <\ga$,
then
\begin{equation}
\label{eq:trace_perturb_carleman}
\left| d_r(S+\Delta,\ga) -d_r(S,\ga) \right|
\leq  \frac{\ga^{-1} \carlnorm{\Delta}}{ 1 - \gamma^{-1} \carlnorm{\Delta} }  \eqsp,
\quad \quad \text{for} \quad r=1,2 \eqsp .
\end{equation}
If $d_2(S,\ga) \hsnorm{S^{-1/2} \Delta S^{-1/2}} <1$, then
\begin{align}
\label{eq:trace_perturb_hs_d1}
\left|  d_1(S + \Delta,\ga) - d_1(S,\ga)  \right|
 &\leq \frac{d_2(S,\ga) \hsnorm{ S^{-1/2} \Delta S^{-1/2}}}{1- d_2(S,\ga) \hsnorm{ S^{-1/2} \Delta S^{-1/2} }}  \eqsp, \\
\label{eq:trace_perturb_hs_d2}
\left|  d_2(S + \Delta,\ga) - d_2(S,\ga)  \right|
 &\leq \frac{ \hsnorm{ S^{-1/2} \Delta S^{-1/2}}}{1-  \hsnorm{ S^{-1/2} \Delta S^{-1/2} }}  \eqsp .
\end{align}
\end{lem}

\begin{proof}
If $\norm{\left((S + \gamma \Id)^{-1}\Delta \right\}} < 1$,
then we may write
\begin{align*}
(S + \Delta + \gamma \Id)^{-1}(S+\Delta) &=(\Id + (S +\gamma \Id)^{-1}\Delta)^{-1} (S +\gamma \Id)^{-1}(S+\Delta)\\
&= \sum_{k=0}^{\infty} (-1)^{k} \left\{(S +\gamma \Id)^{-1}\Delta \right\}^k (S +\gamma \Id)^{-1}(S+\Delta)\\
&= (S +  \gamma \Id)^{-1} S + \sum_{k=1}^{\infty} (-1)^{k} \left\{(S + \gamma \Id^{-1}) \Delta \right\}^k
\left((S + \gamma \Id)^{-1} S - \Id\right)\eqsp ,
\end{align*}
where the series converge in operator-norm.
Since the trace is continuous in the space of trace-class operators,
and using $\openorm{ (S + \gamma \Id)^{-1} S - \Id } <1$,
we get by linearity of the trace,
\begin{multline}
\label{eq:trace_expansion}
\left| d_1(S+\Delta,\gamma) - d_1(S,\gamma) \right| =
\left| \trace{\left\{(S + \Delta + \gamma \Id)^{-1}(S+\Delta) \right\}}
- \trace{\left\{(S + \gamma \Id)^{-1} S \right\}} \right| \\
 = \sum_{k=1}^{\infty} \left| \trace{ \left\{ \left\{(S + \gamma \Id)^{-1} \Delta \right\}^k
 \left\{ (S + \gamma \Id)^{-1} S - \Id \right\} \right\} } \right|
 \leq \sum_{k=1}^{\infty} \left|
 \trace{ \left\{ \left((S + \gamma \Id)^{-1} \Delta \right)^k  \right\} } \right| \eqsp .
\end{multline}
Applying Lemma~\ref{lem:eric-s-trick} with
$B=\Delta $, and $A= (S + \gamma \Id)^{-1}$, we obtain
\begin{align*}
 \trace{ \left\{ \left((S + \gamma \Id)^{-1} \Delta \right)^k  \right\} }
&= \sum_{p=1}^{\infty} \pscal{e_p}{\left((S + \gamma \Id)^{-1} \Delta \right)^k e_p}\\
&= \sum_{p_1=1}^{\infty} \dots \sum_{p_k=1}^{\infty}
\left\{ \left(\prod_{j=1}^k (\la_{p_j} + \gamma)^{-1} \right) \left(\prod_{j=1}^k
 \pscal{e_{p_j}}{ \Delta e_{p_{j+1}}}\right) \right\} \eqsp .
\end{align*}
Since, for all $1 \leq j \leq k$, we have
$\left|  \pscal{e_{p_j}}{ \Delta e_{p_{j+1}}} \right| \leq \norm{\Delta e_{p_j}}$ and $(\la_{p_j} + \gamma)^{-1} \leq \gamma^{-1}$,
the upper-bound in (\ref{eq:trace_expansion})
is actually the sum of a geometric series whose ratio is
$\gamma^{-1} \sum_{p=1}^{\infty} \norm{\Delta e_p}
= \gamma^{-1}\carlnorm{\Delta}$, where $\gamma^{-1}\carlnorm{\Delta}<1$
by assumption, which completes the proof
of (\ref{eq:trace_perturb_carleman}) when $r=1$.
A similar reasoning as above allows to prove (\ref{eq:trace_perturb_carleman}) when $r=2$.

We now prove the second upper-bound (\ref{eq:trace_perturb_hs_d1}).
Using that
\[ \left|  \trace{ \left\{ \left((S + \gamma \Id)^{-1} \Delta \right)^k  \right\} } \right|
 =  \left|  \trace{ \left[ \left\{ \left( S^{1/2}(S + \gamma \Id)^{-1}S^{1/2} \right) \left( S^{-1/2}\Delta S^{-1/2} \right) \right\}^k  \right] } \right| \eqsp,
\]
we may apply Lemma~\ref{lem:eric-s-trick} again, but with
$B= S^{-1/2}\Delta S^{-1/2} $, and $A= S^{1/2}(S + \gamma \Id)^{-1}S^{1/2}$, yielding
\begin{multline*}
 \trace{ \left\{ \left((S + \gamma \Id)^{-1} \Delta \right)^k  \right\} }
= \sum_{p=1}^{\infty} \pscal{e_p}{\left((S + \gamma \Id)^{-1} \Delta \right)^k e_p}\\
= \sum_{p_1=1}^{\infty} \dots \sum_{p_k=1}^{\infty}
\left\{ \left(\prod_{j=1}^k (\la_{p_j} + \gamma)^{-1}\la_{p_j} \right) \left(\prod_{j=1}^k
 \pscal{e_{p_j}}{\left(S^{-1/2}\Delta S^{-1/2}\right)  e_{p_{j+1}}}\right) \right\} \eqsp .
 \end{multline*}
 Then, using that $\left|  \pscal{e_{p_j}}{\left(S^{-1/2}\Delta S^{-1/2}\right)e_{p_{j+1}}} \right|
  \leq \norm{\left(S^{-1/2}\Delta S^{-1/2}\right) e_{p_j}}$,
 and applying H\"{o}lder inequality,
we obtain
\begin{eqnarray*}
&& \left|  \trace{ \left\{ \left((S + \gamma \Id)^{-1} \Delta \right)^k  \right\} } \right|  \\
&\leq& \left\{ \sum_{p=1}^{\infty}  (\la_{p} + \gamma)^{-2}\la_{p}^2 \right\}^{k/2}
\left\{ \sum_{p_1=1}^{\infty} \dots \sum_{p_k=1}^{\infty}
\left(\prod_{j=1}^k
 \pscal{e_{p_j}}{\left(S^{-1/2}\Delta S^{-1/2}\right)  e_{p_{j+1}}}^2 \right)
\right\}^{1/2} \\
&\leq& d^k(S) \thinspace \hsnorm{ S^{-1/2} \Delta S^{-1/2} }^{k} \eqsp .
\end{eqnarray*}
Finally, going back to (\ref{eq:trace_expansion}), the upper-bound
is actually the sum of a geometric series whose ratio is
$d(S) \hsnorm{ S^{-1/2} \Delta S^{-1/2} }$, where $d(S) \hsnorm{ S^{-1/2} \Delta S^{-1/2} }<1$
by assumption, which completes the proof
of (\ref{eq:trace_perturb_hs_d1}).
As for (\ref{eq:trace_perturb_hs_d2}),  observe that
\begin{align*}
\left|d_{2}(S+\Delta, \ga)-d_{2}(S, \ga)\right|
&\leq
 \sum_{k=1}^{\infty}  \hsnorm{ \left\{ \left\{(S + \gamma \Id)^{-1} \Delta \right\}^k
 \left\{ (S + \gamma \Id)^{-1} S - \Id \right\} \right\} } \\
 &\leq
  \sum_{k=1}^{\infty}  \hsnorm{  \left\{(S + \gamma \Id)^{-1} \Delta \right\}^k} \\
  &\leq
    \sum_{k=1}^{\infty}  \hsnorm{  \left\{ S^{-1/2} \Delta S^{-1/2} \right\}}^k
   \eqsp ,\\
\end{align*}
where we used the inequality $\hsnorm{AB} \leq \hsnorm{A} \hsnorm{B}$,
and $\openorm{(S + \gamma \Id)^{-1} S - \Id} \leq 1$
and $\openorm{(S + \gamma \Id)^{-1} S} \leq 1$.
\end{proof}
%%%%%%%%%%%%%%%%%%%%%%%%%%%%%%%%%%%%%%%%%%%%%%%%%%%%%%%%%%%%%%%%%%%%%%

%%%%%%%%%%%%%%%%%%%%%%%%%%%%%%%%%%%%%%%%%%%%%%%%%%%%%%%%%%%%%%%%%%%%%%
\section{Miscellaneous proofs}

\begin{prop}
\label{prop:approx_limit_dist_null_gaf}
Assume (A\ref{hypA:bounded_kernel}) and (B\ref{hypB:sq_sum_eigs}).
Assume in addition that $\P_1 = \P_2 = \P$. If \condgaf~, then
\begin{equation}
\text{Sup}_{x}
\left|\P(T_{\infty}(\hS_W,\ga) \leq x) - \P(T_{\infty}(\S_W,\ga) \leq x) \right| \to 0 \eqsp ,
\end{equation}
where $T_{\infty}(S,\ga)$ for a trace-class operator $S$ is defined in (\ref{eq:definition-limiting-distribution}).
\end{prop}

\begin{proof}
First, define the random variables $\{ Y_n\}$ and $\{ Y \}$ as follows
\begin{align*}
Y_n \eqdef \sump (\hla_p + \ga)^{-1}\hla_p (Z_p^2 -1) \eqsp , \quad
Y   \eqdef \sump ( \la_p + \ga)^{-1} \la_p (Z_p^2 -1) \eqsp ,
\end{align*}
where $\{Z_p\}_{p\geq 1}$ are independent standard normal variables.
Considering the random element $h \in \H$, such that $\pscal{h}{e_p}_{\H}=Z_p$ for all $p \geq 1$,
we may write
\begin{align*}
Y_n &= \norm{(\hS_W + \ga \id)^{-1/2}\hS_W^{-1/2} h}_{\H}^2 - d_{1,n}(\hS_W, \ga) \eqsp , \\
Y    &= \norm{ (\S_W + \ga \id)^{-1/2}\S_W^{-1/2}  h}_{\H}^2 - d_{1,n}( \S_W, \ga) \eqsp .
\end{align*}
Then, using  Eq.~\eqref{eq:trace_perturb_carleman} for $r=1$
in Lemma~\ref{lem:perturb_cov} with $S = \S_W $, and Corollary \ref{cor:hs_conv_cov_op} which gives $\hsnorm{\hS_W - \S_W} = O_{P}(n^{-1/2})$, we get $|Y_n  - Y| = O_P (n^{-1/2})$,
and hence that $Y_n \plim Y$ in case \condgaf~.
Next, applying the Polya theorem~\citep[Theorem 11.2.9]{Lehmann:Romano:2005} gives the result
\begin{equation*}
 \text{Sup}_{x} \left|\P(Y_n \leq x) - \P(Y \leq x) \right| \to 0 \eqsp .
\end{equation*}
\end{proof}
%%%%%%%%%%%%%%%%%%%%%%%%%%%%%%%%%%%%%%%%%%%%%%%%%%%%%%%%%%%%%%%%%%%%%%%%%%%%%%%%%%%%%%%%%%%%%%%%%%%%%%%%%%%%%%%%%%%%%%%%%%%%%%%%%%%%%%%%%

%%%%%%%%%%%%%%%%%%%%%%%%%%%%%%%%%%%%%%%%%%%%%%%%%%%%%%%%%%%%%%%%%%%%%%%%%%%%%%%%%%%%%%%%%%%%%%%%%%%%%%%%%%%%%%%%%%%%%%%%%%%%%%%%%%%%%%%%%
\section{Eigenvalues of covariance operators}
\label{sec:widom}
In this section, we give new general results regarding the decay of eigenvalues of covariance operators. We assume that we have a bounded density $p(x)$ on $\Rset^p$ with respect to the Lebesgue measure, and a translation invariant kernel $k(x-y)$ with positive integrable Fourier transform. In this section, we consider eigenvalues of the second order moment operator, which dominates the covariance operator.
From the proof of Proposition~\ref{prop:range}, the eigenvalues of the  second order moment operator are the eigenvalues
of the following operator from $\ltwo(\Rset^p)$ to $\ltwo(\Rset^p)$, defined as
$$Q f(x)= \int_{\Rset^p} p(x)^{1/2} k(x-y) f(y) p(y)^{1/2} dy
$$
We let denote $\lambda_n(p,K)$ the  eigenvalues of this operator ranked in decreasing order.

We let denote $T(p)$ the pointwise multiplication by $p$, defined from $L^2(\rb^p)$ to $L^2(\rb^p)$. We also denote $C(k)$ the convolution operator by $k$. We thus get $Q = T(p)^{1/2} C(k) T(p)^{1/2}$. Note that by taking Fourier transforms ($P$ of $p$, and $K$ of $k$), the eigenvalues are the same as the one of $ T(K)^{1/2} C(P) T(K)^{1/2} $ and thus $p$ and $K$ plays equivalent roles~\citep{Widom:1963}.

 The following lemma, taken from~\citet{Widom:1964}, gives an upperbound of the eigenvalues in the situation where $p$ and $K$ are indicator functions:
\begin{lem}
Let $\varepsilon>0$. Then there exists $\delta>0$ such that,
if $p(x)$ is the indicator function of $[-1,1]$ and $K$ is the indicator function of $[-\gamma,\gamma]$, with $\gamma \leqslant (1-\varepsilon) n \pi  / 2$, then $\lambda_n(p,K) \leqslant e^{- n \delta}$.
\end{lem}
This result is very useful because it is uniform in $\gamma$, as  long as $\gamma \leqslant (1-\varepsilon) n \pi  / 2$. We now take $\varepsilon=\frac{1}{2}$, and we thus get
$\lambda_n(1_{[-1,1]} , 1_{[-n\pi/4,n\pi/4]} ) \leqslant e^{- n \delta}$ for some $\delta>0$.

We consider the tail behavior of $p(x)$ and of the Fourier transform $K(\omega)$ of $k$, through
$M(p,u) = \max_{ \| x \|_\infty \geqslant u} p(x) $ and
$M(K,v) = \max_{ \| \omega \|_\infty \geqslant v} K(\omega) $, where, for $x=(x_1,\dots,x_p)$, $ \|x\|_\infty = \max_{1 \leq i \leq p}|x_i|$.
 We also let denote $M_0(K)$ and $M_0(p)$ the supremum of $K$ and $p$ over $\rb^d$.

\begin{prop}
For all $(u,v)$ such that $uv = n \pi/4$, then
$$ \lambda_n(p,K) \leqslant M(p,u)M_0(K) + M(K,v) M_0(p)  + M_0(K) M_0(p) e^{-\delta n^{1/p} } $$
\end{prop}
\begin{proof}
We divide twice $\Rset^p$ in two parts, the spatial version $\Rset^p
= \{ x, \ \|x\|_\infty \leqslant u \} \cup
\{ x, \ \|x\|_\infty >  u \}  = A_u \cup B_u $
and the Fourier version $\Rset^p
= \{ \omega , \ \|\omega\|_\infty \leqslant v \} \cup
\{ \omega , \ \|\omega\|_\infty > v \} = A_v \cup B_v$.
We have for all $p$ and $K$,
$$\lambda_n(p,K) \leqslant \lambda_n(p 1_{A_u},K) + \lambda_1( p 1_{B_u},K) $$
which is classical results for perturbation of eigenvalues
By definition of
$M(p,u) $, we have $T(p 1_{B_u}) \preccurlyeq M(p,u) \Id$, and moreover $C(k) \preccurlyeq M_0(K) \Id $, which implies
that $\lambda_1( p 1_{B_u},K) \leqslant M(p,u) M_0(K)$. We thus get
$$\lambda_n(p,K) \leqslant \lambda_n(p 1_{A_u},K) +M(p,u) M_0(K). $$
Similarly, we get
$$\lambda_n(p,K) \leqslant \lambda_n(p 1_{A_u},K 1_{A_v} ) +M(p,u) M_0(K)
+M(K,v) M_0(p).
$$
We know that if two operators satisties $A \preccurlyeq B$, then $\lambda_n(A) \leqslant \lambda_n(B)$, thus since
$T(p 1_{A_u}) \preccurlyeq  T( M_0(p) 1_{A{u}})$ and similarly for $K$, we get
$$\lambda_n(p,K) \leqslant M_0(K) M_0(p)  \lambda_n( 1_{A_u}, 1_{A_v} ) +M(p,u) M_0(K)
+M(K,v) M_0(p)
$$
By a simple change of variable, it easy to show that
$ \lambda_n( 1_{A_u}, 1_{A_v} ) =
\lambda_n( 1_{A_1}, 1_{A_{vu}} )
$
When $p=1$, we immediately have $\lambda_n ( 1_{A_1}, 1_{A_{vu}} ) \leqslant e^{-\delta n}$. When $p>1$, then we notice that the eigenfunctions and eigenvalue of the operators will be product of
eigenfunctions and eigenvectors of the univariate operators. That is, the eigenvalues are of the form $ \mu_{i_1} \cdots \mu_{i_p} $
where $(i_1,\dots,i_p)$ are positive integer and $\mu_i \leqslant e^{-\delta i}$ are eigenvalues of the univariate operator. From the product formulation, we get that if $n$ is equal to the number  of partitions of a certain integer $k$ into $p$ strictly
positive integers, then $\lambda_n \leqslant e^{ -\delta k}$. This number of partitions is exactly equal to
$ [(p-1)! (p-k)!]^{-1} (k-1)! \leqslant (k-1)^p $.

Thus, given any $n$, we can find an integer $k$ such that $(k-1)^p \leqslant n$, and we have $\lambda_n ( 1_{A_1}, 1_{A_{vu}} ) \leqslant e^{-\delta k}$. This leads to $\lambda_n ( 1_{A_1}, 1_{A_{vu}}) \leqslant e^{-\delta n^{1/p} }$. The proposition follows.
\end{proof}

We can now derive a number of corollaries:
\begin{cor}
If $p(x)$ is upper bounded by a constant times $e^{- \alpha \|x\|^2}$ and $K(\omega)$ is upper bounded by a constant times $e^{- \beta \|\omega \|^2}$, then there exists $\eta>0$ such that
$\lambda_n(p,K) = O(  e^{-\eta n^{1/p}} )$.
\end{cor}
\begin{proof}
Take $ u = v = \sqrt{ n \pi / 4}$.
\end{proof}

\begin{cor}
If $p(x)$ is upper bounded by a constant times $( 1 + \|x\| )^{-\alpha}$ (with $\alpha > p$ such that we have integrability) and $K(\omega)$ is upper bounded by a constant times $e^{- \beta \|\omega \|^2}$, then
$\lambda_n(p,K) = O( \frac{1 }{ n^{\alpha-\eta} } )$ for any $\eta>0$.
\end{cor}
\begin{proof}
Take $ v $ proportional to $n^{\eta/\alpha} $.
\end{proof}

\begin{cor}
If $p(x)$ is upper bounded by a constant times $( 1 + \|x\| )^{-\alpha}$ (with $\alpha > p$ such that we have integrability) and $K(\omega)$ is upper bounded by a constant times $( 1 + \|x\| )^{-\beta}$ (with $\beta > p$ such that we have integrability) , then
$\lambda_n(p,K) = O( n^{ -\alpha\beta/(\alpha+\beta) } )$.
\end{cor}
\begin{proof}
Take $ v $ proportional to $  n^{ \alpha/(\alpha+\beta) }   $.
\end{proof}
%%%%%%%%%%%%%%%%%%%%%%%%%%%%%%%%%%%%%%%%%%%%%%%%%%%%%%%%%%%%%%%%%%%%%%%%%%%%%%%%%%%%%%%%%%%%%%%%%%%%%%%%%%%%%%%%%%%%%%%%%%%%%%%%%%%%%%%%%

\bibliography{ktest.bib}
\end{document}